%% file: main.tex
\documentclass[10pt,twocolumn,letterpaper]{article}

\usepackage{cvpr}
\usepackage{times}
\usepackage{epsfig}
\usepackage{graphicx}
\usepackage{amsmath}
\usepackage{amssymb}
\usepackage{booktabs}
\usepackage[dvipsnames]{xcolor}
\usepackage{caption}
\usepackage{appendix}
\usepackage{subfigure}
\usepackage[pagebackref=true,breaklinks=true,letterpaper=true,colorlinks,bookmarks=false]{hyperref}

\definecolor{mycolor}{RGB}{198,88,124}

\newcommand{\myparagraph}[1]{\vspace{0.1em}\noindent\textbf{#1}}

\newcommand\blfootnote[1]{%
  \begingroup
  \renewcommand\thefootnote{}\footnote{#1}%
  \addtocounter{footnote}{-1}%
  \endgroup
}

\cvprfinalcopy 

\def\cvprPaperID{} 

\ifcvprfinal\fi

\begin{document}

\title{Generating 3D People in Scenes without People}

\author{
  Yan Zhang\textsuperscript{*}$^{1,3}$,\; Mohamed Hassan$^2$,\; Heiko Neumann$^3$,\; Michael J. Black$^2$,\; Siyu Tang\textsuperscript{*}$^{1}$ \\
  $^1$ ETH Z\"{u}rich, Switzerland \\
  $^2$Max Planck Institute for Intelligent Systems, T\"{u}bingen, Germany \\
  $^3$Institute of Neural Information Processing, Ulm University, Germany
}

\twocolumn[{%
\renewcommand\twocolumn[1][]{#1}%
\maketitle
\begin{center}
   \newcommand{\teaserwidth}{\textwidth}
   \vspace{-0.5cm}
   \centerline{\includegraphics[width=\linewidth]{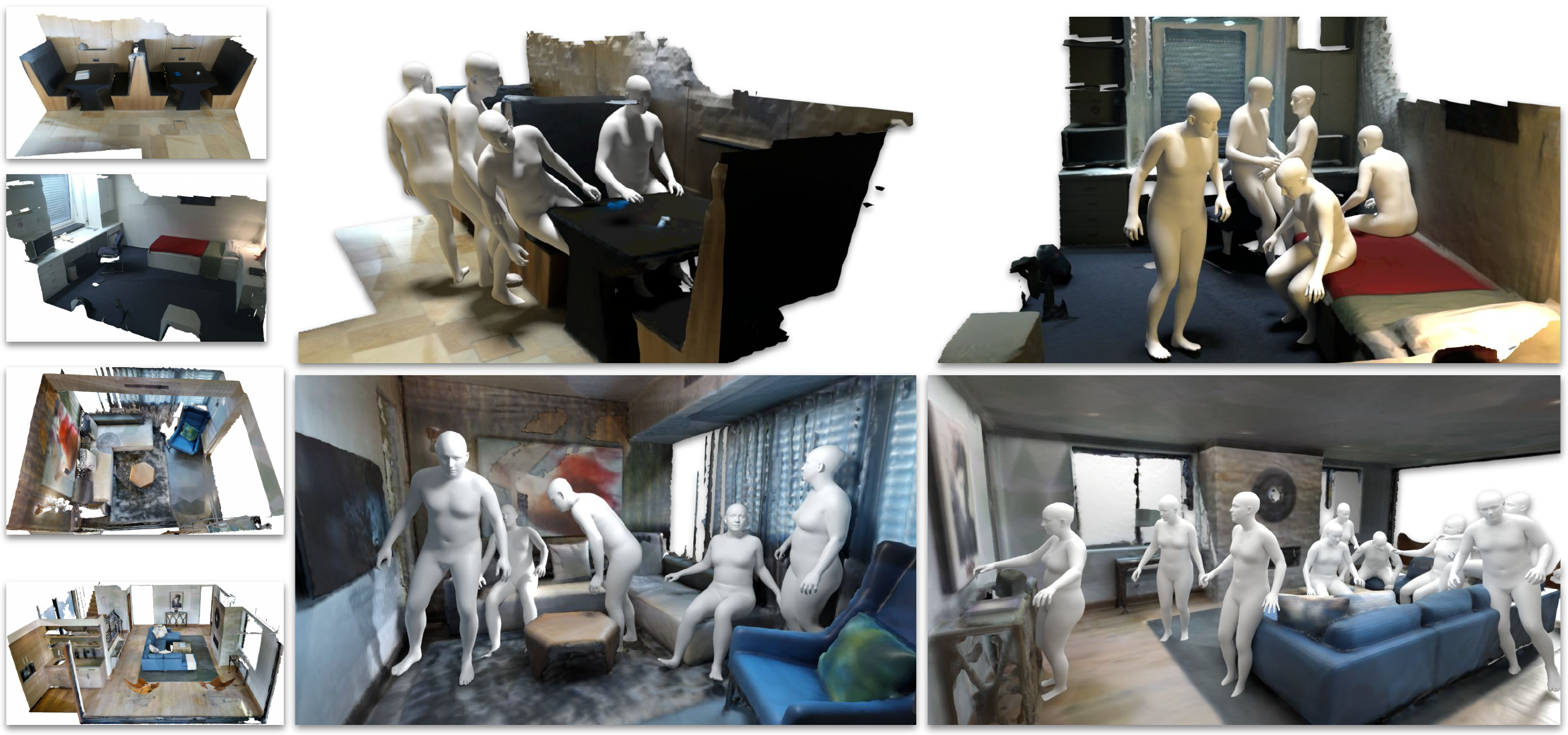}}
    \captionof{figure}{3D human bodies with various shapes and poses are automatically generated to interact with the scene. Appropriate human-scene contact is encouraged, and human-scene surface interpenetration is discouraged.}
    \label{fig:teaser}
\end{center}%
}]

\maketitle
\thispagestyle{empty}

\begin{abstract}
  We present a fully automatic system that takes a 3D scene and generates plausible 3D human bodies that are posed naturally in that 3D scene. 
  Given a 3D scene without people, humans can easily imagine how people could interact with the scene and the objects in it. 
  However, this is a challenging task for a computer as solving it requires that (1) the generated human bodies to be semantically plausible within the 3D environment (e.g. people sitting on the sofa or cooking near the stove), and (2) the generated human-scene interaction to be physically feasible such that the human body and scene do not interpenetrate while, at the same time, body-scene contact supports physical interactions. 
  To that end, we make use of the surface-based 3D human model SMPL-X.
  We first train a conditional variational autoencoder to predict semantically plausible 3D human poses conditioned on latent scene representations, then we further refine the generated 3D bodies using scene constraints to enforce feasible physical interaction. 
  We show that our approach is able to synthesize realistic and expressive 3D human bodies that naturally interact with 3D environment. 
  We perform extensive experiments demonstrating that our generative framework compares favorably with existing methods, both qualitatively and quantitatively. We believe that our scene-conditioned 3D human generation pipeline will be useful for numerous applications; e.g.~to generate training data for human pose estimation, in video games and in VR/AR. Our project page for data and code can be seen at: \url{https://vlg.inf.ethz.ch/projects/PSI/}.
  \blfootnote{$^*$ This work was performed when Y. Z. and S. T. were at MPI-IS and University of T\"{u}bingen.}
\end{abstract}


\input{latex/sec1-intro.tex}

\input{latex/sec2-related_work.tex}
\input{latex/sec3-method.tex}
\input{latex/sec4-experiment.tex}

\input{latex/sec5-conclusion.tex}

{\small
\bibliographystyle{ieee_fullname}
\bibliography{egbib}
}


\clearpage

\begingroup
\onecolumn 

\appendix
\renewcommand{\thefigure}{S\arabic{figure}}
\renewcommand{\thetable}{S\arabic{table}}

\setcounter{figure}{0} 
\setcounter{table}{0} 
\section*{Appendix}

\section{Experiment Details}
\subsection{From PROX-Qualitative to {\bf PROX-E}}
The PROX-Qualitative (or {\bf PROX-Q} for short) dataset comprises recordings of 20 subjects in 12 indoor scenes, including 3 bedrooms, 5 living rooms, 2 sitting booths and 2 offices.
The 3D scenes were scanned with a commercial Structure Sensor RGB-D camera and reconstructed by the accompanying 3D reconstruction solution Skanect.
We refer to \cite{PROX:2019} for more details of how {\bf PROX-Q} was created.
Note that the scene meshes of {\bf PROX-Q} do not form valid rooms, i.e. there is no ceiling and some walls are missing. Furthermore, the meshes are not semantically segmented.

To our knowledge, {\bf PROX-Q} is the largest dataset capturing real human-scene interactions at the 3D mesh level. However, due to the incomplete room scans and lack of mesh semantics, we extend {\bf PROX-Q} as described below to serve our purposes of human-scene interaction modeling and generation from the viewpoint of an embodied agent: 

\paragraph{(1) Building up virtual walls, floors, ceilings.} To achieve this goal, we import the scene meshes of {\bf PROX-Q} into Blender, which we use to enclose the original scene meshes to create rooms. With complete rooms, when we  image the 3D scene using a virtual camera, we can always obtain a completed depth map. 
The completed depth maps are illustrated in Fig.~\ref{fig:prox_e}.

\paragraph{(2) Semantic annotation of the scene meshes.} The mesh semantics follow the Matterport3D dataset \cite{Matterport3D}, which incorporates 40 categories of common indoor objects\footnote{One can see the object categorization via:  \url{https://github.com/niessner/Matterport/blob/master/metadata/mpcat40.tsv}}. Our annotation is performed manually, and the mesh vertex color denotes the object labels.
Our virtual images of the scene, therefore capture both the scene depth and semantics.

\paragraph{(3) Setting up virtual cameras.} The original {\bf PROX-Q} dataset only incorporates video recordings from a single view in each scene; this gives only have 12 depth-semantics pairs to use for training. This is too limited to learn generalization to new scenes. 
To overcome this, for each individual frame captured by the real camera, we create a set of virtual cameras in the scene to capture the human behavior. The virtual cameras are posed according to the room structure and the human body position. Specifically, we create a 3D grid according to the room size. The range of width and length is determined by the size of the room.
The range of height is between the pelvis of the human body and the ceiling height that we have created. For each camera, the X-axis is parallel to the ground, and the Z-axis is towards the human body center. Next, we add Gaussian noise on the camera translations, and discard views with no human bodies or strong body occlusions; i.e., we keep views where the body part around the pelvis ($\pm$ 10 pixels) is not occluded by any object in the scene. We argue that such noise is essential. Otherwise the generated human bodies will always be located in the center of the depth-semantic maps. Furthermore, we only keep the virtual cameras with the distance to the human body between 1.65m and 6.5m, so that the projected body sizes to the virtual cameras are similar to the body sizes captured by real cameras. Fig.~\ref{fig:noise} shows a set of virtual cameras before and after applying the Gaussian noise to the camera translations. Moreover, the resolution of depth and semantics is set to 480$\times$270, and the camera intrinsic parameters are
\begin{equation}
    K = \begin{pmatrix}
    233.826 & 0 & 239.5 \\
    0 & 233.826 & 134.5 \\
    0 & 0 & 1 
    \end{pmatrix},
\end{equation}
which is a default setting in Open3D \cite{Zhou2018} after specifying the depth/semantics resolution.

\begin{figure}
    \centering
    \includegraphics[width=\linewidth]{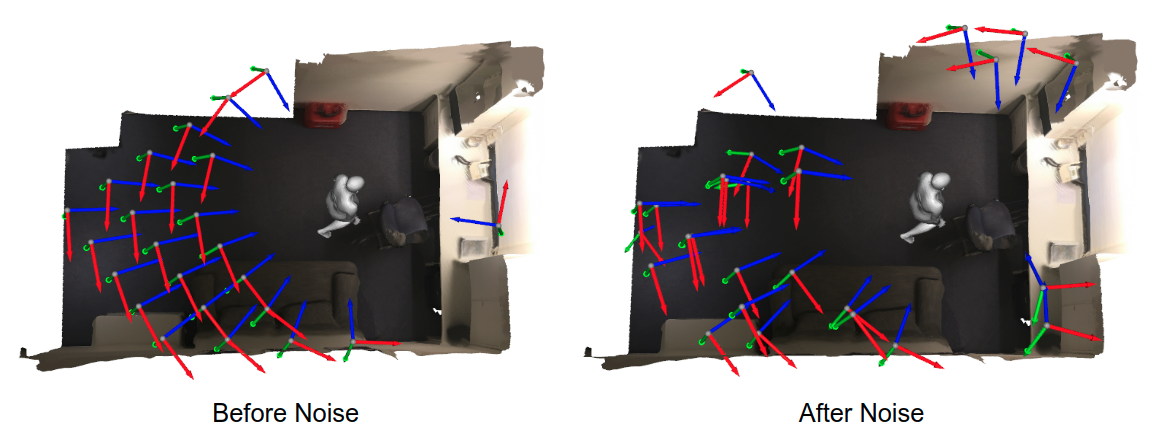}
    \caption{Illustration of the virtual cameras before and after applying the Gaussian noise to the camera translation. The X, Y and Z axes of each camera are denoted by red, green and blue, respectively. }
    \label{fig:noise}
\end{figure}

\begin{table}[t]
    \centering
    \caption{The seven rooms in {\bf MP3D-R}, retrieved from the Matterport3D dataset \cite{Matterport3D}.}
    \begin{tabular}{ccc}
    \toprule
    scan ID  & region ID & room type \\
    \midrule
    17DRP5sb8fy & 0-0 & bedroom \\
    17DRP5sb8fy & 0-8 & family room \\
    17DRP5sb8fy & 0-7 & living room \\
    sKLMLpTHeUy & 0-1 & family room \\
    X7HyMhZNoso & 0-16 & living room \\
    zsNo4HB9uLZ & 0-0 & bedroom \\
    zsNo4HB9uLZ & 0-13 & living room\\
    \bottomrule
    \end{tabular}
    \label{tab:app_mp3dr}
\end{table} 

\subsection{Creating the MP3D-R dataset}
Our {\bf MP3D-R} dataset is extracted from the Matterport3D dataset \cite{Matterport3D}. 
We extract the 7 rooms by annotating bounding boxes of regions, as shown in Tab. \ref{tab:app_mp3dr}. These 7 rooms have room types that are similar to {\bf PROX-E}.

When trimming the rooms according to the annotation, we expand the annotated bounding box size by 0.5 meters to ensure that walls, ceilings and floors are incorporated. 
Note that, the Habitat simulator and the original Matterport3D dataset have different gravity directions. The Habitat simulator assumes that the gravity direction is along $-Y$. Thus, after loading the scene meshes from Matterport3D, we rotate the scene mesh by $-90$ degree w.r.t.~the X-axis to match the bounding box annotation from Habitat. Fig.~\ref{fig:mp3dr_appendix} shows some retrieved room meshes with the world coordinate origins.

\begin{figure}
    \centering
    \includegraphics[width=\linewidth]{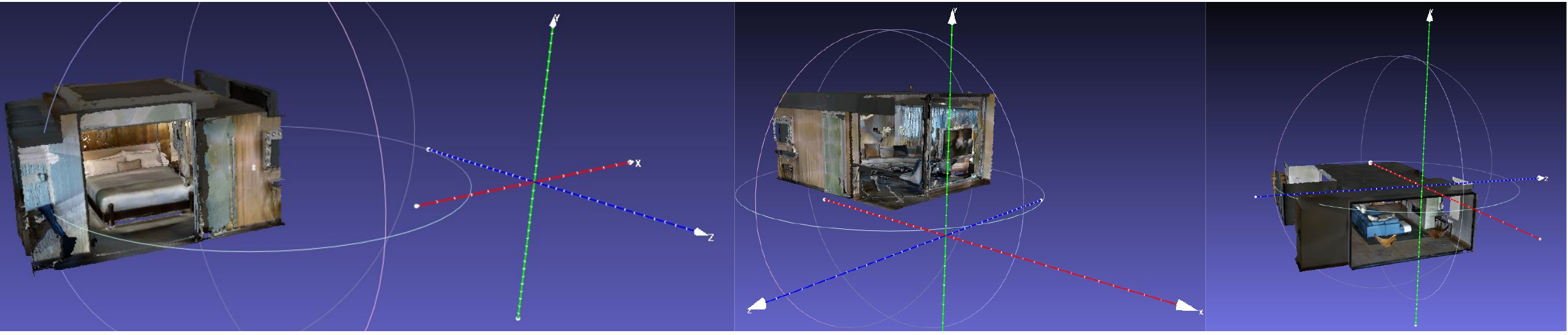}
    \caption{Three examples of the rooms in {\bf MP3D-R}. One can see the world origins, and the gravity direction is along $-Y$.}
    \label{fig:mp3dr_appendix}
\end{figure}

Next, we use the Habitat simulator \cite{habitat19iccv} to create a virtual agent in the room. In each scene, we first put the agent in the room center, and then manipulate that virtual agent to cruise around the room. According to ranges of virtual cameras in {\bf PROX-E}, we set the height of agent sensor to 1.8 meters from the ground. For each snapshot, we record the RGB image, the scene depth, the scene semantics, as well as the camera extrinsic parameters. The frame resolution and the camera intrinsics are identical to our settings in {\bf PROX-E}.

Following the pipeline for creating {\bf PROX-Q}, we also compute scene signed distance functions (SDFs) of the {\bf MP3D-R} scenes. For each room, we first use Poisson surface reconstruction to convert the meshes to be watertight. Fig.~\ref{fig:app_wt} shows an example of the reconstructed scene mesh. Next, similar to \cite[Sec. 3.6]{PROX:2019} we compute the SDF in a uniform voxel grid of size $256 \times 256 \times 256$ which spans a padded bounding box of the scene mesh.

\begin{figure}
    \centering
    \includegraphics[width=\linewidth]{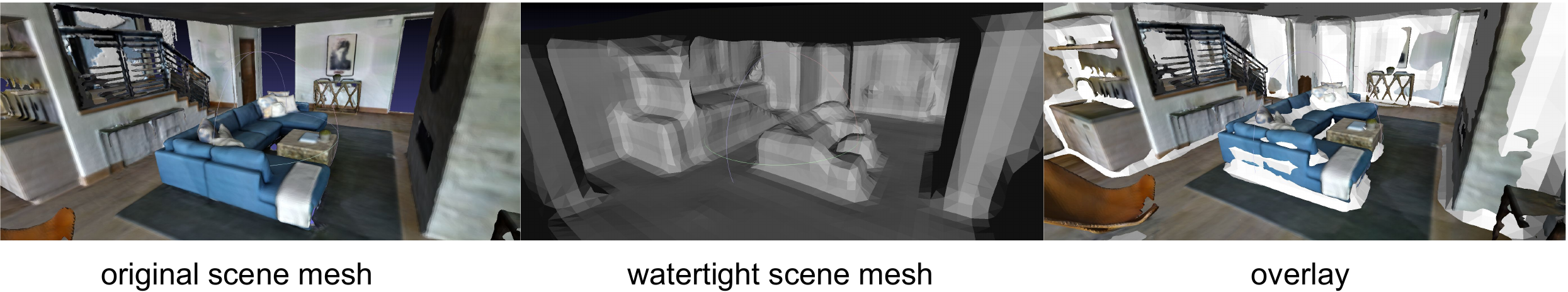}
    \caption{From left to right: The original scene mesh, the mesh after Poisson surface reconstruction, and their overlay.}
    \label{fig:app_wt}
\end{figure}

\begin{figure}[t]
    \centering
    \includegraphics[width=\linewidth]{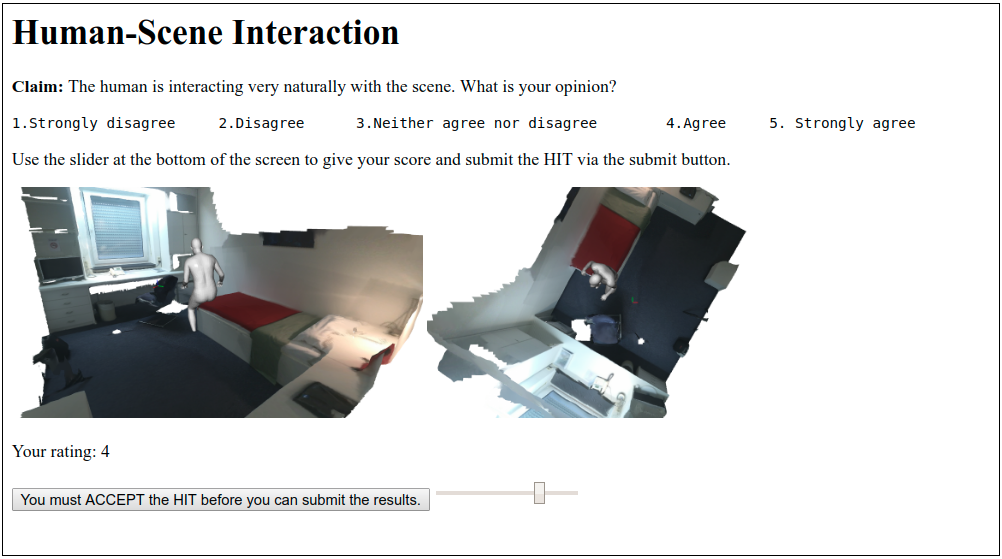}
    \caption{The user interface of our user study, in which the users are requested to rate how naturally the human is interacting with the environment.}
    \label{fig:app_userstudy}
\end{figure}

\subsection{Details of the baseline method}

To our knowledge, the most related work is Li et al. \cite{li2019putting}, which aims to put humans in a scene and infer the affordances of 3D indoor environments. The authors first propose an efficient and fully-automatic 3D human pose synthesizer to generate stick figures, using a pose prior learned from a large-scale 2D dataset \cite{wang2017binge} and the physical constraints from the target 3D scenes. With this pose synthesizer, the authors create a dataset incorporating synthesized human-scene interactions. Next, based on the synthesized dataset, the authors develop a generative model for 3D affordance prediction, which is able to generate body stick figures based on the scene images.

Compared to the method of Li et al. \cite{li2019putting}, our solution has the following key differences: (1) Our {\bf PROX-E} dataset contains {\em real} human-scene interactions rather than synthesized ones. This is highly beneficial to model the distribution of human-scene interactions in the real world. (2) Our solution is to generate body meshes rather than 3D body stick figures (See Fig.~5 in \cite{li2019putting}). Therefore, the results can be directly used in applications like VR, AR and others. (3) We use the SMPL-X model \cite{SMPL-X:2019} in our work, hence our methods can generate various body shapes and fine-grained hand poses, beyond the body global configurations and local poses. (4) SMPL-X can be regarded as a differentable function mapping from human body features to human body meshes, so the physical constraints applied on the body mesh surfaces can be back-propagated to the body features like in \cite{PROX:2019}. 
(5) We use scene depth and semantics to represent the scene, rather than using RGB (or RGBD) images as in \cite{li2019putting}. In our study, the RGB images are only available from the limited number of real camera views in {\bf PROX-E}, and hence using RGB images can increase the risk of overfitting. In addition, the benefits of scene depth and semantics are revealed in \cite{zhou2019does}.

Therefore, in our work, we modify the method of Li at al. \cite{li2019putting} as mentioned in Sec. \ref{sec:baseline}, so that their model can generate body meshes like our method, and a fair comparison can be conducted. We treat the modified version of \cite{li2019putting} as our baseline. We train the baseline model with the {\bf PROX-E} dataset like training our models. After generating body meshes in test scenes, we also apply our scene geometry-aware fitting to refine the results of the baseline model. 
The qualitative results of the baseline with fitting are shown in Fig.~\ref{fig:app-proxe-baseline} and Fig.~\ref{fig:app_mp3d_baseline}. We argue that our modification is necessary and favorable to the baseline to produce high quality 3D human bodies. For the quantitative comparison, please refer to Tab.~\ref{tab:res_gen_diversity}, Tab.~\ref{tab:res_gen_physics} and Tab.~\ref{tab:user_study}, in the main paper.

\subsection{Details of the user study}

To evaluate how naturally the human body meshes are posed in the scenes, we perform a user study via Amazon Mechanical Turk (AMT). For each generated body pose, we render images of the body-and-scene mesh from two different views. Fig.~\ref{fig:app_userstudy} shows our AMT user interface. We pose the hypothesis that the human body interacts with the scene in a very natural manner, and then ask subjects to judge the correctness of this hypothesis. Their judgements are recorded on a 5-point Likert scale.

Unlike the user study in \cite{li2019putting,wang2017binge}, we do not show pairs of results from different methods in the user interface. Instead, we have multiple methods to compare, and hence let the Turker  evaluate each individual result in order to keep the user interface clear. Also, we report the standard errors of the user study results in addition to the mean values, which indicate how reliable the  scores are. One can see in Tab.~\ref{tab:user_study} that the scene geometry-aware fitting can reduce the the standard error, indicating that subjects tend to give more consistent judgements. The ground truth has the lowest standard error, which indicates that Turkers are able to judge when the human-scene interaction is natural.

\subsection{More discussions on model training}
We discussed loss weights and training schemes in Sec.~3.4. The weights are determined empirically: 
First, the KL-divergence weight is 0.1 for better representation power of the latent variables, as indicated in \cite{burgess2018understanding,higgins2017beta}.
The annealing scheme effectively avoids a collapsed VAE posterior, which outputs a constant result no matter how the latent variable varies. 
Second, the VPoser weight 0.001 is determined referring to \cite{PROX:2019,SMPL-X:2019}, to balance plausibility and variability of generated body poses. 
A too small weight increases body pose variations but can lead to implausible body poses (e.g. twisted legs).
Additionally, in our trials the $\mathcal{L}_{HS}$ weights are set to avoid overfitting. Also, enabling $\mathcal{L}_{HS}$ earlier during training causes bad body reconstruction, since the modified Chamfer loss in Eq. \eqref{eq:contact} can pull the preliminary reconstructed body mesh to the closest scene mesh vertices. 
Moreover, the scene geometry-aware fitting loss weights are larger than the training loss weights, so as to reduce number of iterations in optimization while retaining the quality.

\section{Generative Model Latent Space Analysis}
We show how the body smoothly changes in Fig. \ref{fig:vis_z_2}. Note that the results are \textit{without} scene geometry-aware fitting.
First, Fig. \ref{fig:vis_z_2} indicates that our model effectively learns natural human-scene interactions. It shows that the generated body tends to stand when located on the floor, touches the desk (Fig. \ref{fig:vis_z_2} (2)), and sits on the bed (Fig. \ref{fig:vis_z_2} (4)) when located close to the furniture. 
Second, the body configurations are disentangled in the latent space to some extent.
For example, the body in Fig. \ref{fig:vis_z_2} (2) mainly moves along the X direction in the world coordinate system, while the body in Fig. \ref{fig:vis_z_2} (3) mainly moves along the Y direction. 
Third, the body pose is less plausible when its latent variable is far away from zero. This is similar to VPoser \cite{SMPL-X:2019}; Tab.~5 in the manuscript shows the benefits of our model used as a scene-dependent pose prior.

\begin{figure}
    \centering
    \includegraphics[width=\linewidth]{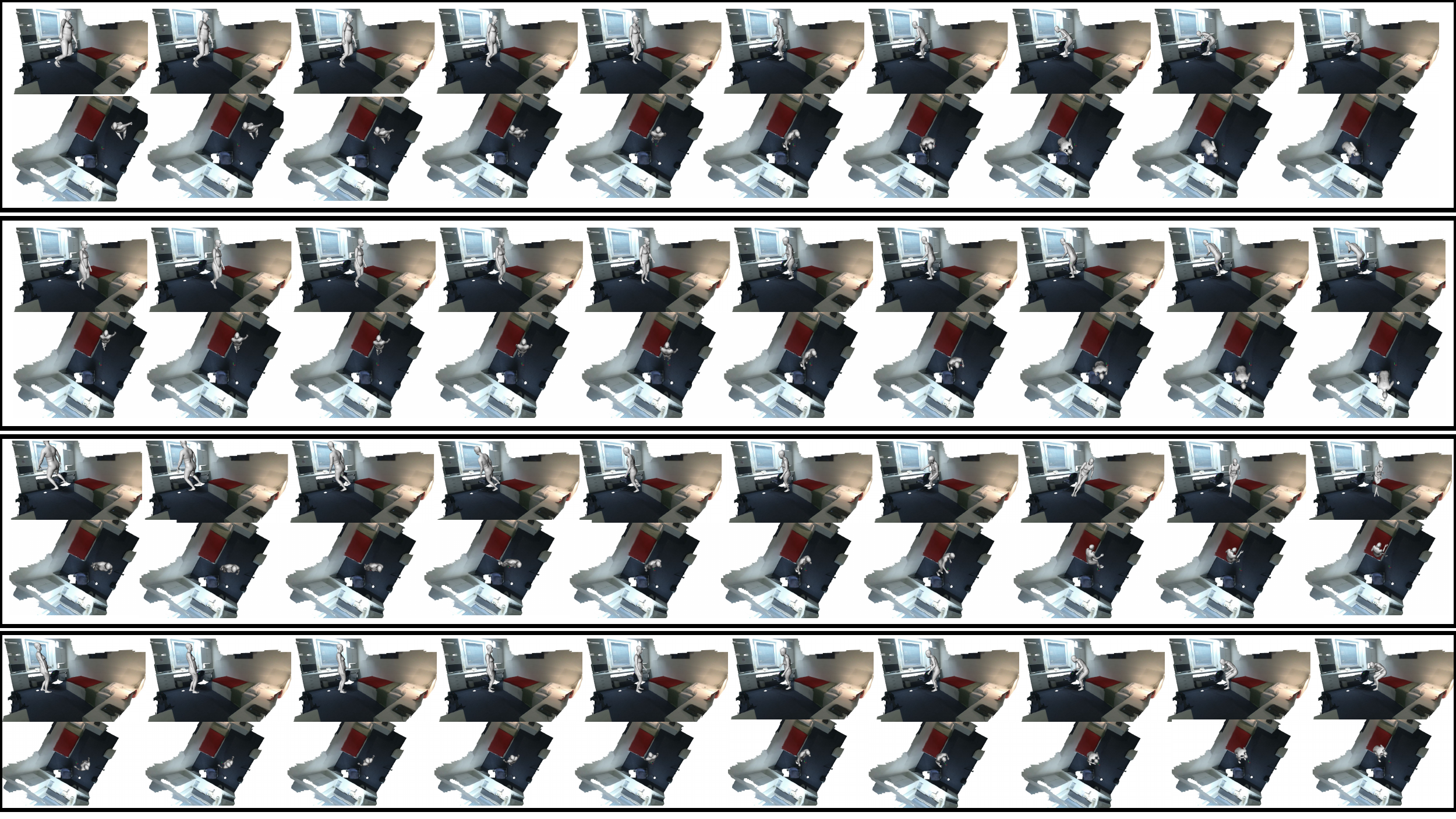}
    \caption{Illustration of the 32D latent space in the one-stage model. We regularly sample points along a line ranging from -3 to 3, and show the body meshes from two views. From top to bottom: (1) All dimensions of the line change. (2) The first 16 dimensions change, and the rest are zero. (3) Only the last 16 dimensions change. (4) Only the middle 16 dimensions change.}
    \label{fig:vis_z_2}
\end{figure}

\section{More Qualitative Results and Failure Cases}

Fig.~\ref{fig:app-proxe-baseline} and Fig.~\ref{fig:app-proxe-ours} show qualitative results in the test scenes of {\bf PROX-E}. Fig.~\ref{fig:app_mp3d_baseline} and Fig.~\ref{fig:app_mp3d_ours} show qualitative results of the generative models and the scene geometry-aware fitting in {\bf MP3D-R}.

We find that failure cases can be categorized to two cases: 
First, the generative model is not always reliable in test scenes, since samples from the model are not always plausible. 
Some results sampled from the generative model cannot match the geometric structures in the test scenes, and hence the body floats in the air, or collides with the scene mesh. See Fig.~\ref{fig:app-fail1} for examples. Such failure cases can occur in both the baseline and our methods. 
Second, although the scene geometry-aware fitting can effectively resolve floating and collision, its optimization process cannot simulate all real physics such as gravity and elasticity. Therefore, it could hurt the human-scene interaction semantics of the results produced by the generative model. Fig.~\ref{fig:app-fail2} shows some examples of such failure cases, which contain abnormal body global configurations and human-scene contact caused by our scene geometry-aware fitting.

Moreover, we discover that the quality of generated bodies also depends on the test scene data quality. For example, we have observed many low-quality generations in {\bf MP3D-R}, which has more complex scene structures and noisy scans (unknown surfaces floating in the air) than {\bf PROX-E}. Noisy scans can lead to noisy depth and semantic segmentation, and complex geometric structures can make geometry-aware fitting fail.

\section{Details of Scene-aware 3D Body Pose Estimation}
In the experiment presented in Sec.~\ref{sec:pose_estimation}, we follow the work of \cite{PROX:2019}, and modify its Eq. (1) to incorporate our learned scene-dependent pose prior. The equation (1) in \cite{PROX:2019} is given by
\begin{equation}
    \begin{split}
        E(\beta, \theta, \varphi, \gamma, M_s) &= E_J + \lambda_D E_D + \lambda_{\theta_b} E_{\theta_b} + \lambda_{\theta_f} E_{\theta_f} + \lambda_{\theta_h} E_{\theta_h} \\ 
        &+ \lambda_{\alpha} E_{\alpha} + \lambda_{\beta} E_{\beta} + \lambda_{\epsilon} E_{\epsilon} + \lambda_{\epsilon} E_{\epsilon} \\
        &+ \lambda_{P} E_{P} + + \lambda_{C} E_{C},
    \end{split}
\end{equation}
where the notation corresponds to that in \cite{PROX:2019}. In our work, we only modify the VPoser regularizer, i.e., $E_{\theta_b} = \|\theta_b\|_2^2$, and leave the other terms unchanged. Specifically, we change it to 
\begin{equation}
    E_{\theta_b} = \|\theta_b - \theta_b^s\|_2^2,
\end{equation}
in which $\theta_b^s$ is our scene-dependent pose prior. We have demonstrated how to derive the $\theta_b^s$ in Sec. \ref{sec:pose_estimation}. During optimization, the initial pose feature is set to $\theta_b^s$, instead of a zero vector as in \cite{PROX:2019}. In our trials, changing the weight to $1.5\lambda_{\theta_b}$ yields  better performance.

\begin{figure}
    \centering
    \includegraphics[width=\linewidth]{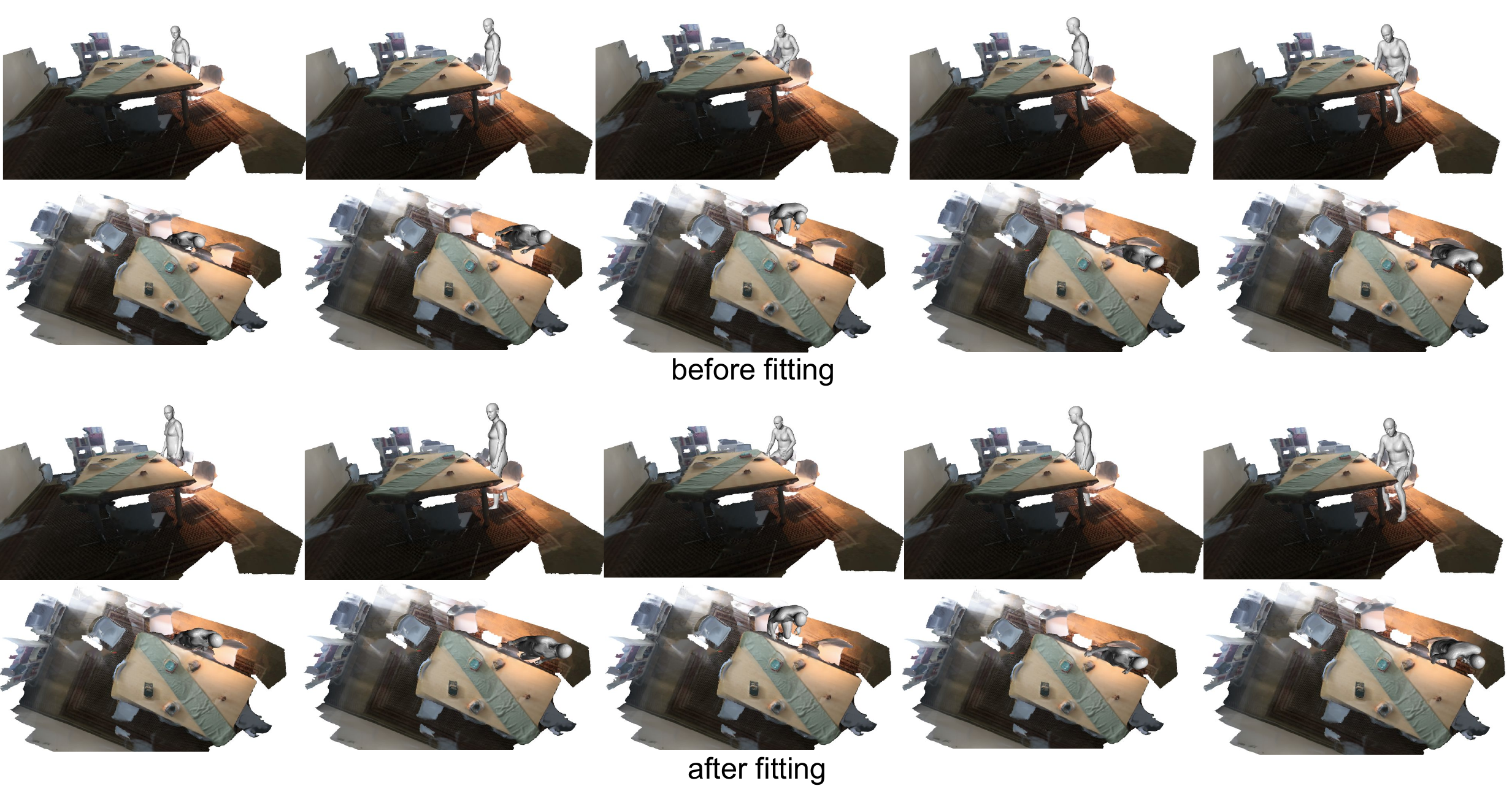}
    \vspace{2cm}
    \includegraphics[width=\linewidth]{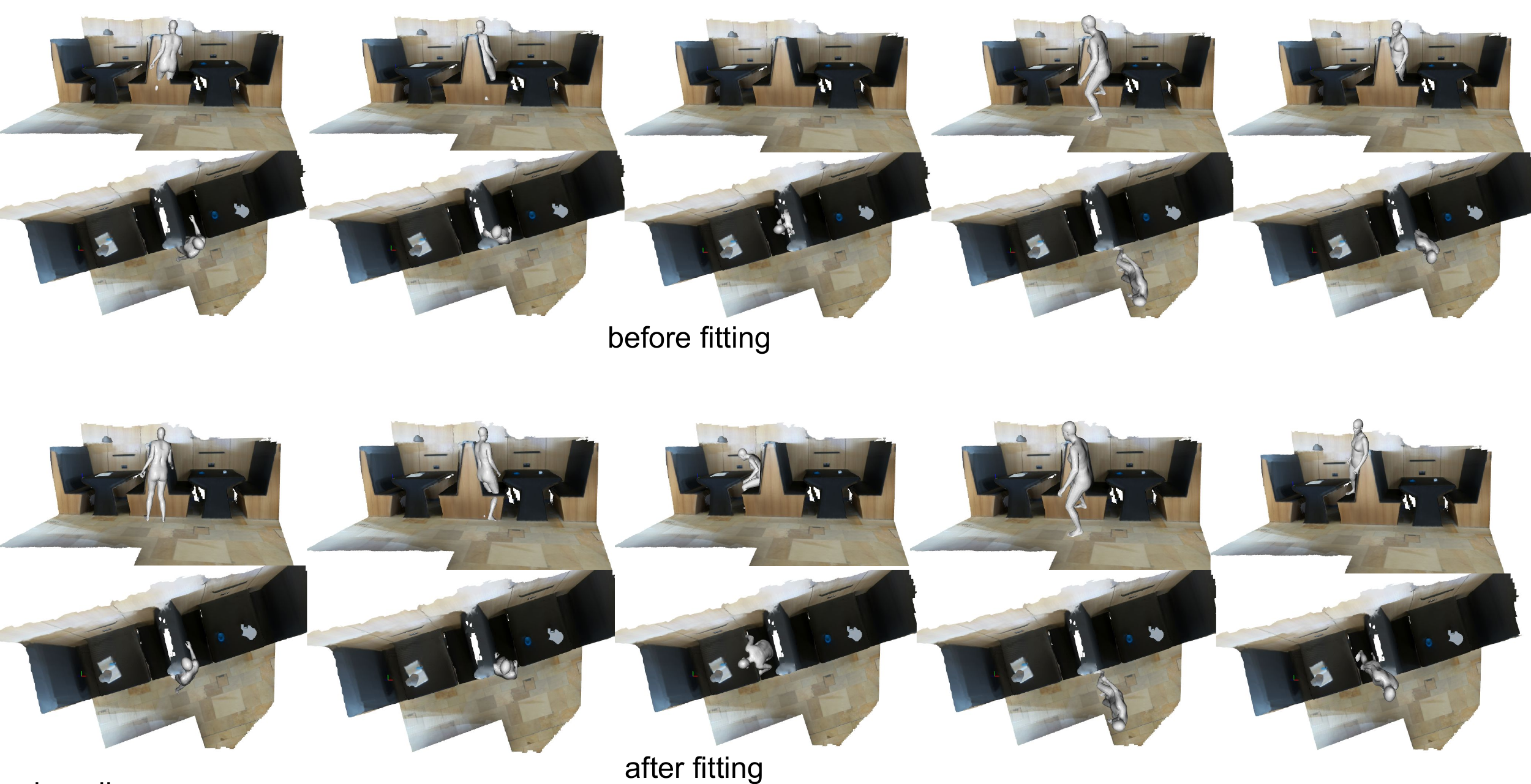}
    \caption{Qualitative results of the \textcolor{blue}{ baseline} method in {\bf PROX-E}. 
    The results before and after the scene geometry-aware fitting are shown.}
    \label{fig:app-proxe-baseline}
\end{figure}

\begin{figure}
    \centering
    \includegraphics[width=\linewidth]{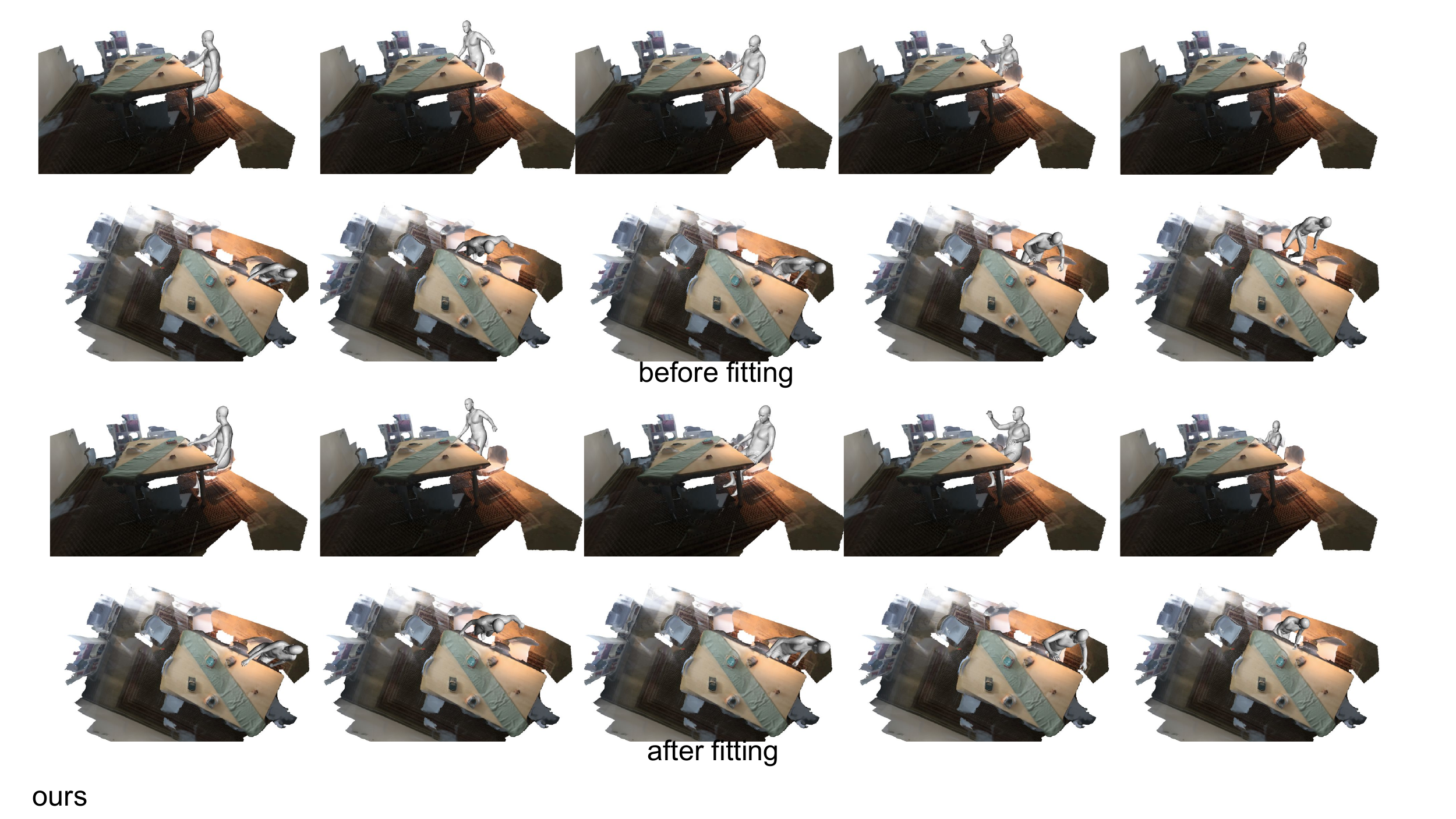}
    \vspace{2cm}
    \includegraphics[width=\linewidth]{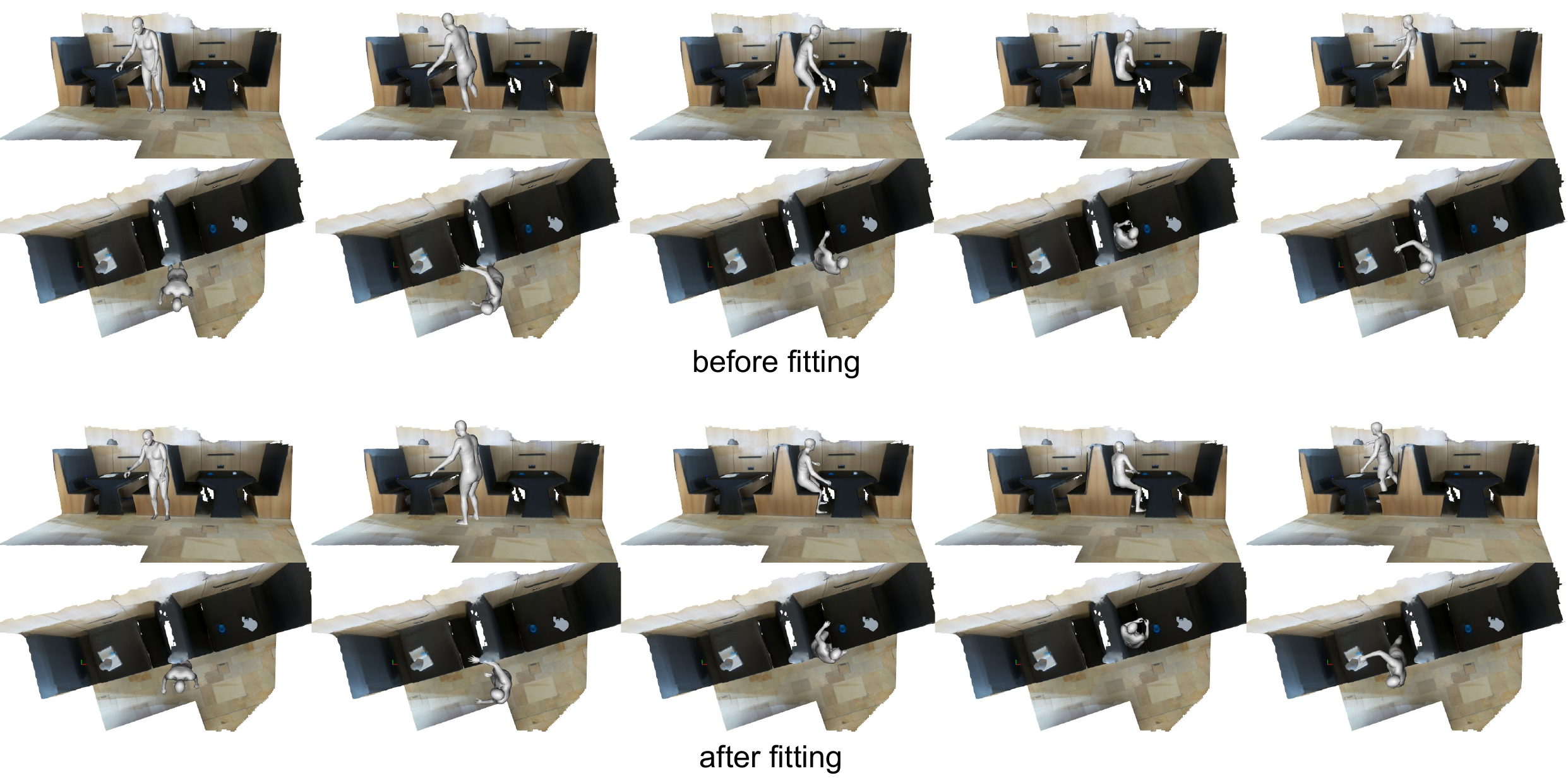}
    \caption{Qualitative results of  \textcolor{blue}{S1} in  {\bf PROX-E}. The results before and after the scene geometry-aware fitting are shown.}
    \label{fig:app-proxe-ours}
\end{figure}

\begin{figure}[h!]
    \centering
    \includegraphics[width=\linewidth]{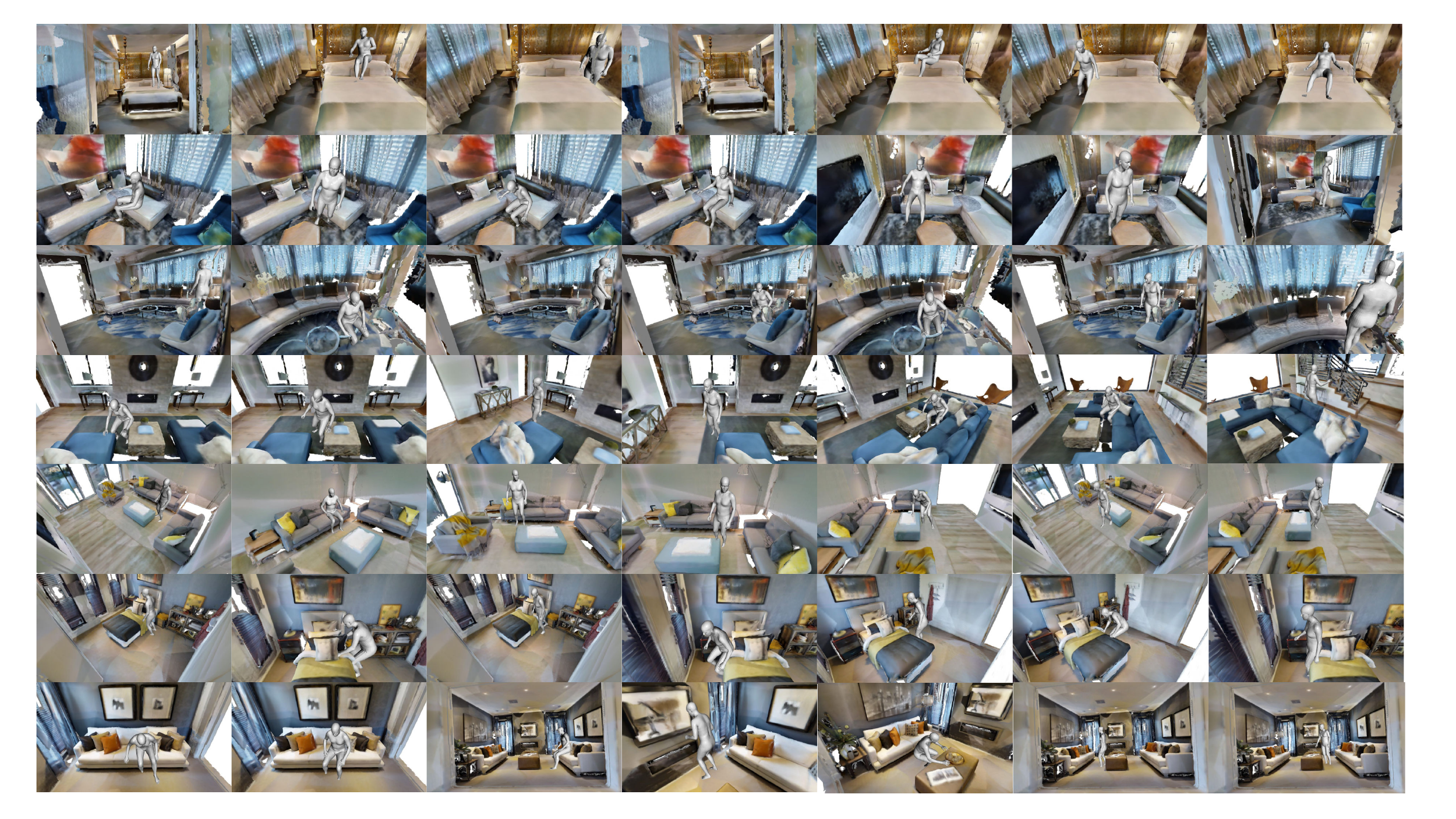}
    \caption{Qualitative results of the \textcolor{blue}{baseline} with fitting in {\bf MP3D-R}.
    We argue that our modifications to \cite{li2019putting} are necessary and favorable to produce high quality 3D human bodies.
     For the quantitative comparison, please refer to Tab.~\ref{tab:res_gen_diversity}, Tab.~\ref{tab:res_gen_physics} and Tab.~\ref{tab:user_study}}
    \label{fig:app_mp3d_baseline}
\end{figure}

\begin{figure}[h!]
    \centering
    \includegraphics[width=\linewidth]{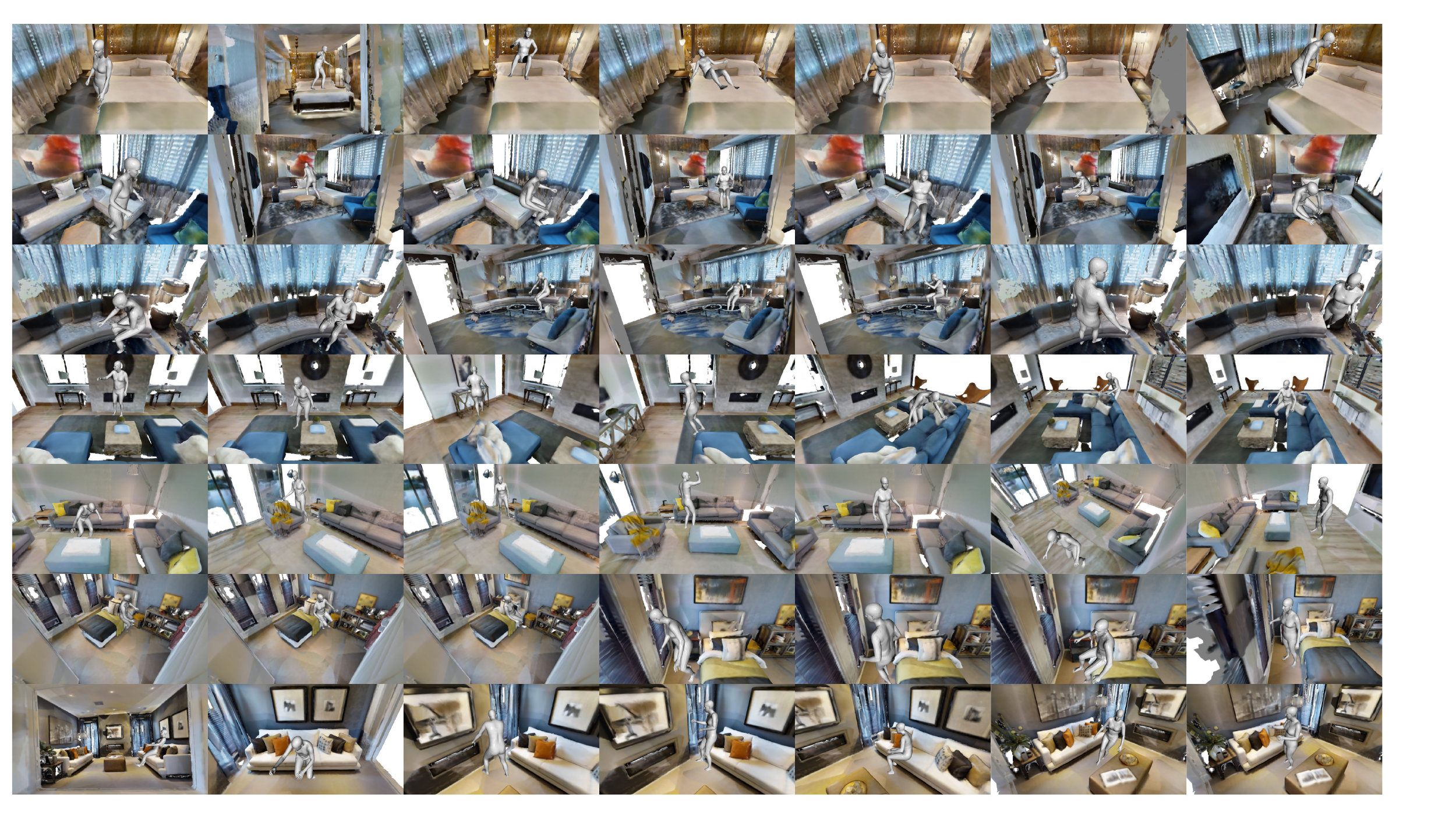}
    \caption{Qualitative results of \textcolor{blue}{ S1} with fitting in {\bf MP3D-R}.}
    \label{fig:app_mp3d_ours}
\end{figure}

\begin{figure}
    \centering
    \includegraphics[width=\linewidth]{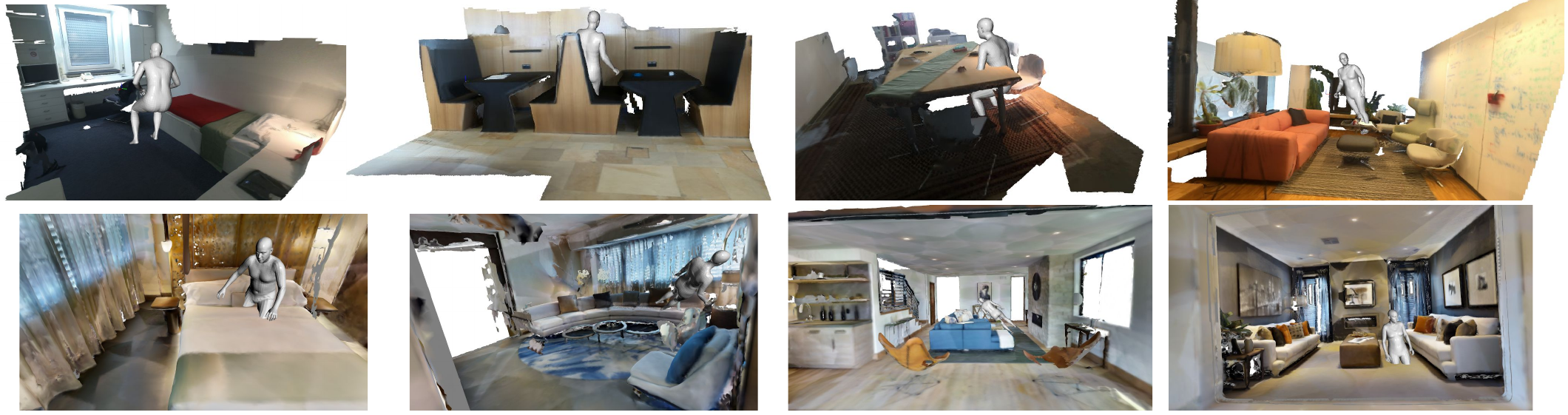}
    \caption{Failure cases of body mesh generation. One can see the body floating and colliding with the scene mesh, which are implausible in the real world.}
    \label{fig:app-fail1}
\end{figure}

\begin{figure}
    \centering
    \includegraphics[width=\linewidth]{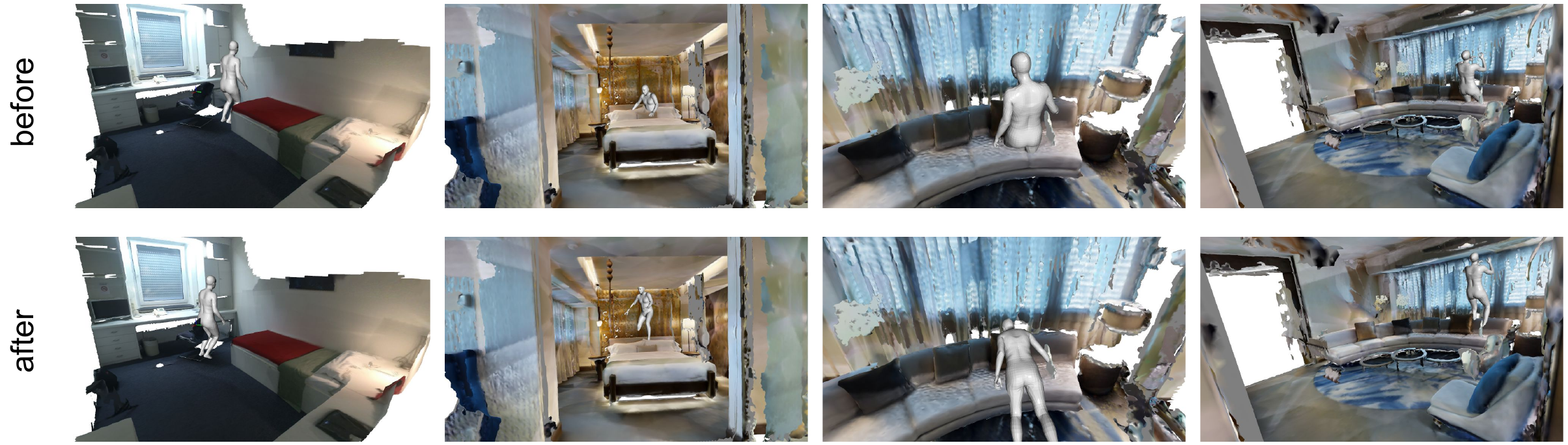}
    \caption{Failure cases of the scene geometry-aware fitting, for which results before and after the fitting are presented. One can see abnormal body translation, rotation and body-scene contact in the real world.}
    \label{fig:app-fail2}
\end{figure}

\end{document}

%% file: latex/sec1-intro.tex
\section{Introduction}

In recent years, many high-quality datasets of 3D indoor scenes have emerged such as Matterport3D \cite{Matterport3D}, Replica \cite{replica19arxiv}, and Gibson \cite{xiazamirhe2018gibsonenv}, which employ 3D scanning and reconstruction technologies to create digital 3D environments. Also, virtual robotic agents exist inside of  3D environments such as Gibson \cite{xiazamirhe2018gibsonenv} and the Habitat simulator \cite{habitat19iccv}.
These are used to develop scene understanding methods from embodied views, thus providing platforms for indoor robot navigation, AR/VR, computer games and many other applications. 
Despite this progress, a significant limitation of these environments is that they do not contain people.
The reason such worlds contain no people is that there are no automated tools to generate realistic people interacting realistically with 3D scenes, and manually doing this requires significant artist effort.
Consequently, our goal is to automatically generate natural and realistic 3D human bodies in the scene. 
The generated human bodies are expected to be physically plausible (e.g.~neither floating nor interpenetrating), diverse, and posed naturally within the scene.
This is a step towards equipping high-quality 3D scenes and simulators (e.g.~Matterport3D \cite{Matterport3D} and Habitat \cite{habitat19iccv}) with semantically and physically plausible 3D humans, and is essential for numerous applications such as creating synthetic datasets, VR/AR, computer games, etc.

Our solution is inspired by how humans infer plausible interactions with the environment.
According to the studies of \cite{you2007applications}, humans tend to propose interaction plans depending on the structure and the semantics of objects. Afterwards, to realize the interaction plan, physical rules will apply to determine the detailed human-object configuration, while guaranteeing that the human body can neither float in the air nor collide into the objects. 
Therefore, our method has two steps: (1) We propose a generative model of human-scene interaction using a conditional variational autoencoder (CVAE) \cite{sohn2015learning} framework. Given scene depth and semantics, we can sample from the CVAE to obtain various human bodies. (2) Next, we transform the generated 3D human body to the world coordinates and  perform scene geometry-aware fitting, so as to refine the human-scene interaction and eliminate physically implausible configurations (e.g.~floating and collision). 

We argue that realistically modeling human-scene interactions requires a realistic model of the body.
Previous studies on scene affordance inference and human body synthesis in the literature, like \cite{li2019putting,wang2017binge,zhu2016inferring}, represent the body as a 3D stick figure or coarse volume.
This prevents detailed reasoning about contact such as how the leg surface contacts the sofa surface. 
Without a model of body shape, it is not clear whether the estimated body poses correspond to plausible human poses.
To overcome these issues,  we use the SMPL-X model \cite{SMPL-X:2019}, which takes a set of low-dimensional body pose and shape parameters and outputs a 3D body mesh with important details like the fingers. 
Since SMPL-X is differentiable, it enables straightforward optimization of human-scene contact and collision prevention \cite{PROX:2019}. In addition, we incorporate the body shape variation in our approach, so that our generated human bodies have various poses and shapes.

To train our method we exploit the  {PROX-Qualitative} dataset \cite{PROX:2019}, which includes 3D people captured moving in 3D scenes.
We extend this by rendering images, scene depth, and semantic segmentation of the scene from many virtual cameras.
We conduct extensive experiments to evaluate the performance of different models for scene-aware 3D body mesh generation. 
For testing, we extract 7 different rooms from the Matterport3D \cite{Matterport3D} dataset and use a virtual agent in the Habitat Simulator \cite{habitat19iccv} to capture scene depth and semantics from different views. 
Based on prior work, e.g.~\cite{li2019putting,wang2017binge}, we propose three metrics to evaluate the diversity, the physical plausibility, and the semantic plausibility of our results. 
The experimental results show that our solution effectively generates 3D body meshes in the scene, and outperforms the modified version of a state-of-the-art body generation method \cite{li2019putting}.
We will make our datasets and evaluation metrics available to establish a benchmark.

Our trained model learns about the ways in which 3D people interact with 3D scenes.
We show how to leverage this in the form of a scene-dependent body pose prior and show how to use this to improve 3D body pose estimation from RGB images.
In summary, our contributions are as follows: (1) We present a solution to generate 3D human bodies in scenes, using a CVAE to generate a body mesh with semantically plausible poses. 
We follow this with scene-geometry-aware fitting to refine the human-scene interaction. (2) We extend and modify two datasets, and propose three evaluation metrics for scene-aware human body generation. We also modify the method of \cite{li2019putting} to generate body meshes as the baseline (see Sec.~\ref{sec:baseline}).
The experimental results show that our method outperforms the baseline. (3) We show that our human-scene interaction prior is able to improve 3D pose estimation from RGB images.

%% file: latex/sec2-related_work.tex
\section{Related work}
Multiple studies focus on placing objects in an image so that they appear natural \cite{dvornik2018importance,lee2018context,lin2018st,ouyang2018pedestrian}. 
For example, \cite{dvornik2018importance,sun2017seeing, torralba2003contextual} use contextual information to predict which objects are likely to appear at a given location in the image.
Lin et al.~\cite{lin2018st} apply homography transformations to 2D objects to approximate the perspectives of the object and background.
Tan et al.~\cite{tan2018and} predict likely positions for people in an input image and retrieve a person that semantically fits into the scene from a database.
Ouyang et al.~\cite{ouyang2018pedestrian} use a GAN framework to synthesize pedestrians in urban scenes. 
Lee et al.~\cite{lee2018context} learn where to place objects or people in a semantic map and then determine the pose and shape of the respective object.
However, all these methods are limited to 2D image compositing or inpainting. 
Furthermore, the methods that add synthetic humans do not take interactions between the humans and world into account.

To model human-object or human-scene interactions it is beneficial to know which interactions are possible with a given object.
Such opportunities for interactions are referred to as affordances \cite{gibson2014ecological} and numerous
works in computer vision have made use of this concept \cite{chuang2018learning,delaitre2012scene,grabner2011makes,gupta20113d,kim2014shape2pose,koppula2014physically,koppula2013learning,li2019putting,savva2016pigraphs,wang2017binge,zhu2014reasoning,zhu2016inferring}.
Object affordance is often represented by a human pose when interacting with a given object \cite{delaitre2012scene,grabner2011makes,gupta20113d,kim2014shape2pose,li2019putting,savva2016pigraphs,wang2017binge,zhu2014reasoning,zhu2016inferring}.
For example, \cite{grabner2011makes, gupta20113d, zhu2016inferring} search for valid positions of human poses in 3D scenes. 
Delataire et al.~\cite{delaitre2012scene} learn associations between objects and human poses in order to improve object recognition. 
Given a 3D model of an object Kim et al.~\cite{kim2014shape2pose} predict human poses interacting with the given object. 
Given an image of an object Zhu et al.~\cite{zhu2014reasoning} learn a knowledge base to predict a likely human pose and a rough relative location of the object with respect to the pose. 
Savva et al.~\cite{savva2016pigraphs} learn a model connecting human poses and arrangement of objects in a 3D scene that can generate snapshots of object interaction given a corpus of 3D objects and a verb-noun pair.
Monszpart et al.~\cite{monszpart2018imapper} use captured human motion to infer the objects in the scene and their arrangement.
Sava et al.~\cite{scenegrok2014savva} predict action heat maps that highlight the likelihood of an action in the scene.
Recently, Chen et al.~\cite{chen2019holistic++} propose to tackle scene parsing and 3D pose estimation jointly and to leverage their coupled nature to improve scene understanding. 
Chao et al.~\cite{chao2019learning} propose to train multiple controllers to imitate simple motions from mocap, and then use hierarchical reinforcement learning (RL) to achieve higher-level interactive tasks.
The work of Zanfir et al. \cite{zanfir2019human} first estimates the ground plane in the image, and requires a foreground person image as input.
The above methods do not use a realistic body model to represent natural and detailed human-environment interactions.

Recently, Wang et al.~\cite{wang2017binge} published an affordance dataset large enough to obtain reliable estimates of the probabilities of poses and to train neural networks on affordance prediction.
The data is collected from multiple sitcoms and contains
images of scenes with and without humans.
The images with humans contain rich behavior of humans interacting with various objects.
Given an image and a location as input, Wang et al.~first predict the most likely pose from a set of 30 poses. 
This pose is deformed and scaled by a second network to fit it into the scene.
Li et al.~\cite{li2019putting} extend this work to automatically estimate where to put people and to predict 3D poses.
To acquire 3D training data they map 2D poses to 3D poses and place them in 3D scenes from the SUNCG dataset \cite{song2017semantic,zhang2017physically}.
This synthesized dataset is cleaned by removal of all predictions intersecting with the 3D scene or without sufficient support for the body.
The methods of \cite{li2019putting,wang2017binge} are limited in their generalization, since they require a large amount of paired data and manual cleaning of the pose detections. 
Such a large amount of data might be hard to acquire for scenes that are less frequently covered by sitcoms, or in the case of \cite{li2019putting} in 3D scene datasets. 
Furthermore, both methods only predict poses represented as {\em stick figures}.
Such a representation is hard to validate visually, lacks details, and can not directly be used to generate realistic synthetic data of humans interacting with an environment.

%% file: latex/sec3-method.tex
\section{Methods}
\label{sec:methods}

\subsection{Preliminaries}

\myparagraph{3D scene representation.}
We represent the scene from the view of an embodied agent, as in the Habitat simulator \cite{habitat19iccv}. 
According to \cite{zhou2019does}, which indicates that the depth and semantic segmentation are the most valuable modalities for scene understanding, we capture scene depth and semantics as our scene representation. 
For each view, we denote the stack of depth and semantics as $x_s$, the camera perspective projection from 3D to 2D as $\pi(\cdot)$, and its inverse operation as $\pi^{-1}(\cdot)$ for 3D recovery. 
Our training data, $x_s$, are generated from Habitat and we resize this to $128 \times 128$ for compatibility with our network; we retain the aspect ratio and pad with zeros where needed. 
The 3D-2D projection $\pi(\cdot)$ normalizes the 3D coordinates to the range of $[-1,1]$, using the camera intrinsics and the maximal depth value. Note that each individual $x_s$ is from a single camera view. We do not use multi-view data in our work.

\myparagraph{3D human body representation.}
We use SMPL-X \cite{SMPL-X:2019} to represent the 3D human body. SMPL-X can be regarded as a function $\mathcal{M}(\cdot)$, mapping a group of low-dimensional body features to a 3D body mesh. The 3D body mesh has 10475 vertices and a fixed topology. In our study, we use the body shape feature $\beta \in \mathbb{R}^{10}$, the body pose feature $\theta_b \in \mathbb{R}^{32}$, and the hand pose feature $\theta_h \in \mathbb{R}^{24}$. The body pose feature $\theta_b$ is represented in the latent space of VPoser \cite{SMPL-X:2019}, which is a variational autoencoder trained on a large-scale motion capture dataset, AMASS \cite{AMASS:2019}. 
The global rotation $R$, i.e., the rotation of the pelvis, is represented by a 6D continuous rotation feature \cite{zhou2019continuity}, which facilitates back-propagation in our trials. The global translation $t$ is represented by a 3D vector in meters. The global rotation and translation is with respect to the camera coordinates. Based on the camera extrinsics, $T_c^w$, one can transform the 3D body mesh to the world coordinates.

We denote the joint body representation as $x_h: = (t, R, \beta, \theta_b, \theta_h)^T \in \mathbb{R}^{75}$; i.e., the concatenation of individual body features. When processing the global and the local features separately as in \cite{li2019putting}, we denote the global translation as $x_h^g$, and the other body features as $x_h^l$.

\subsection{Scene context-aware Human Body Generator}

\subsubsection{Network architecture}
\begin{figure}[t]
    \centering
    \includegraphics[width=\linewidth]{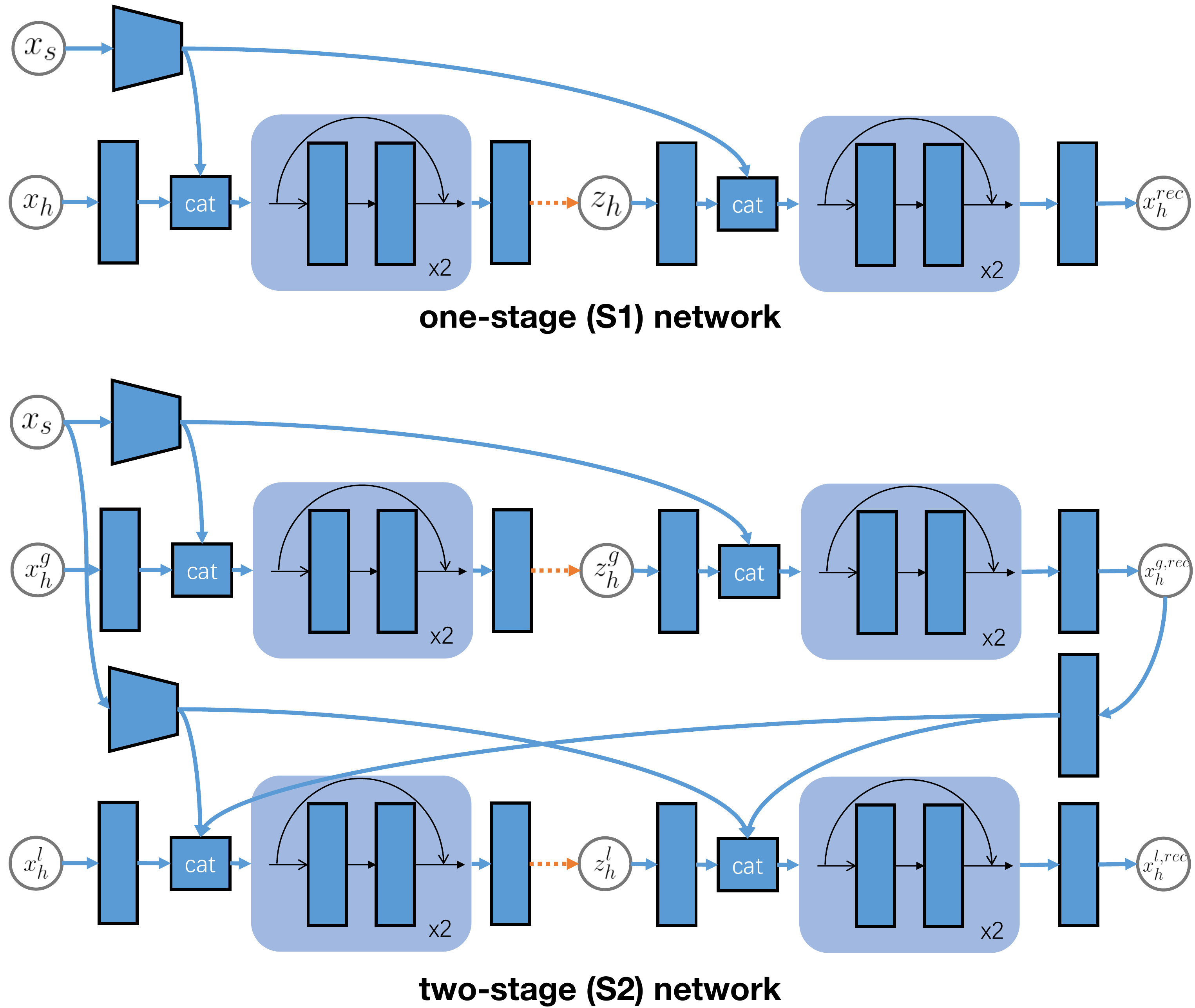}
    \caption{Network diagrams of our models. The trapezoids denote the scene encoders, which are fine-tuned from a pre-trained ResNet18 network. The blue rectangles denote fully-connected (fc) layers. In the residual blocks, Leaky-ReLU \cite{maas2013rectifier} is employed between fc layers. The orange dashed arrows denote the sampling operation in the VAE re-parameterization trick \cite{kingma2013auto}. The blocks with ``cat'' denote the feature concatenation operation.} 
    \label{fig:network_diagram}
\end{figure}

We employ a conditional variational autoencoder (CVAE) \cite{sohn2015learning} framework to model the probability $p(x_h | x_s)$. When inferring all body features jointly, we propose a one-stage (S1) network. When inferring $x_h^g$ and $x_h^l$ successively, we factorize the probability as $p(x_h^l | x_h^g, x_s)p(x_h^g | x_s)$ and use a two-stage (S2) network. 
The network architectures are illustrated in Fig.~\ref{fig:network_diagram}. Referring to \cite{li2019putting}, our scene encoder is fine-tuned from the first 6 convolutional layers in ResNet18 \cite{he2016deep}, which is pre-trained on ImageNet \cite{deng2009imagenet}. The human feature $x_h$ is first lifted to a high dimension (256 in our study) via a fully-connected layer, and then concatenated with the encoded scene feature. In the two-stage model, the two scene encoders are both fine-tuned from ResNet18, but do not share parameters. After the first stage, the reconstructed body global feature $x_h^{g, rec}$ is further encoded, and is used in the second stage to infer the body local features.    

\subsubsection{Training loss}
\label{sec:train_loss}

The entire training loss can be formulated as
\begin{align}
\label{eq:loss_all}
\begin{split}
    \mathcal{L} &= \mathcal{L}_{rec} +\alpha_{kl} \mathcal{L}_{KL} + \alpha_{vp} \mathcal{L}_{\mathit{VPoser}} \\ 
    &+ \alpha_{cont} \mathcal{L}_{contact} + \alpha_{coll} \mathcal{L}_{collision}, 
\end{split}
\end{align}
where the terms denote the reconstruction loss, the Kullback–Leibler divergence, the VPoser loss, the human-scene contact loss and the human-scene collision loss, respectively. The set of $\alpha$'s denotes the loss weights. 
For simplicity, we denote $\alpha_{cont} \mathcal{L}_{contact} + \alpha_{coll} \mathcal{L}_{collision}$ as $\mathcal{L}_{HS}$, implying the loss for human-scene interaction.

\myparagraph{Reconstruction loss $\mathcal{L}_{rec}$:} It is given by $\mathcal{L}_{rec} =$
\begin{equation}
    \label{eq:rec}
     \frac{| x_h^g - x_h^{g, rec} | + | \pi(x_h^g) - \pi(x_h^{g, rec})|}{2} + | x_h^l - x_h^{l, rec} |,
\end{equation}
in which the global translation, the projected and normalized global translation, and the other body features are considered separately. We apply this reconstruction loss in both our S1 model and our S2 model.

\myparagraph{KL-Divergence $\mathcal{L}_{KL}$:} Denoting our VAE encoder as $q(z_h|x_h)$, the KL-divergence loss is given by
\begin{equation}
    \mathcal{L}_{KL} = D_{KL} \left( q(z_h | x_h)~ ||~ \mathcal{N}(0, \mathbf{I}) \right).
\end{equation}
Correspondingly, in our S2 model the KL-divergence loss is given by $\mathcal{L}_{KL} = $
\begin{equation}
    D_{KL} \left( q(z_h^g | x_h^g) || \mathcal{N}(0, \mathbf{I}) \right) + D_{KL} \left( q(z_h^l | x_h^l) || \mathcal{N}(0, \mathbf{I}) \right).
\end{equation}
We use the re-parameterization trick in \cite{kingma2013auto} so that the KL divergence is closed form.

\myparagraph{VPoser loss $\mathcal{L}_{VPoser}$:} Since VPoser \cite{SMPL-X:2019} attempts to encode natural poses with a normal distribution in its latent space, like in \cite{SMPL-X:2019} and \cite{PROX:2019}, we employ the VPoser loss, i.e. 
\begin{equation}
    \mathcal{L}_{VPoser} = | \theta_b^{rec}  |^2, 
\end{equation}
to encourage the generated bodies to have natural poses.

\myparagraph{Collision loss $\mathcal{L}_{collision}$:} Based on the model output $x_h^{rec}$, we generate the body mesh and transform it to world coordinates. Then, we compute the negative signed-distance values at the body mesh vertices given the negative signed distance field (SDF) $\Psi_s^{-}(\cdot)$, and minimize 
\begin{equation}
    \label{eq:collision}
    \mathcal{L}_{coll} = \mathbb{E} \left[  | \Psi_s^{-} \left( T_{c}^w\mathcal{M}(x_h^{rec})  \right) | \right].
\end{equation}
indicating the average of absolute values of negative SDFs on the body.

\myparagraph{Contact loss $\mathcal{L}_{contact}$:} Following \cite{PROX:2019}, we encourage contact between the body mesh and the scene mesh. Hence, the contact loss is written as
\begin{equation}
    \label{eq:contact}
    \mathcal{L}_{contact} = \sum_{v_c \in C\left(T_{c}^w\mathcal{M}(x_h^{rec})\right)    } \min_{v_s \in \mathcal{M}_s} \rho(| v_c - v_s |), 
\end{equation}
in which $C(\cdot)$ denotes selecting the body mesh vertices for contact according to the annotation in \cite{PROX:2019}, $\mathcal{M}_s$ denotes the scene mesh, and $\rho(\cdot)$ denotes the Geman-McClure error function \cite{GemanMcClure1987} for down-weighting the influence of scene vertices far away from the body mesh.

\subsection{Scene geometry-aware Fitting}
\label{sec:fitting}

We refine the body meshes with an optimization step similar to \cite{PROX:2019}. 
It encourages contact and helps to avoid inter-penetration between the  body and the scene surfaces, while not deviating much from the generated pose.
Let the generated human body configuration be $x_h^0$. 
To refine this, we minimize a fitting loss taking into account the scene geometry, i.e.
\begin{align}
\begin{split}
    \mathcal{L}_{f}(x_h) &= |x_h - x_h^0| + \alpha_1 \mathcal{L}_{contact} + \alpha_2 \mathcal{L}_{collision} \\ &+ \alpha_3 \mathcal{L}_{VPoser},
\end{split}
\end{align}
in which the $\alpha$'s denote the loss weights; the loss terms are defined above.

\subsection{Implementation}
Our implementation is based on PyTorch v1.2.0 \cite{paszke2017automatic}. For the Chamfer distance in the contact loss we use the same implementation as \cite{deprelle2019learning, groueix2018b}.
For training, we set $\{ \alpha_{kl}, \alpha_{vp}\}=\{0.1, 0.001\}$ in Eq.~\ref{eq:loss_all} for both our S1 and S2 models, in which $\alpha_{kl}$ increases linearly in an annealing scheme \cite{bowman2015generating}. When additionally using $\mathcal{L}_{HS}$, we set $\{ \alpha_{cont}, \alpha_{coll}\}=\{0.001, 0.01\}$, and enable it after 75\% training epochs to improve the interaction modeling. We use the Adam optimizer \cite{kingma2014adam} with the learning rate $3e^{-4}$, and terminate training after 30 epochs. For the scene geometry-aware fitting, we set $\{ \alpha_{1}, \alpha_{2}, \alpha_{3}\}=\{0.1, 0.5, 0.01\}$ in all cases. 
Our data, code and models will be available for research purposes.

%% file: latex/sec4-experiment.tex
\section{Experiments \protect\footnote{Please see appendix for more details.}}

\subsection{Scene-aware 3D Body Mesh Generation}

\subsubsection{Datasets}
\begin{figure}[t]
    \centering
    \includegraphics[width=\linewidth]{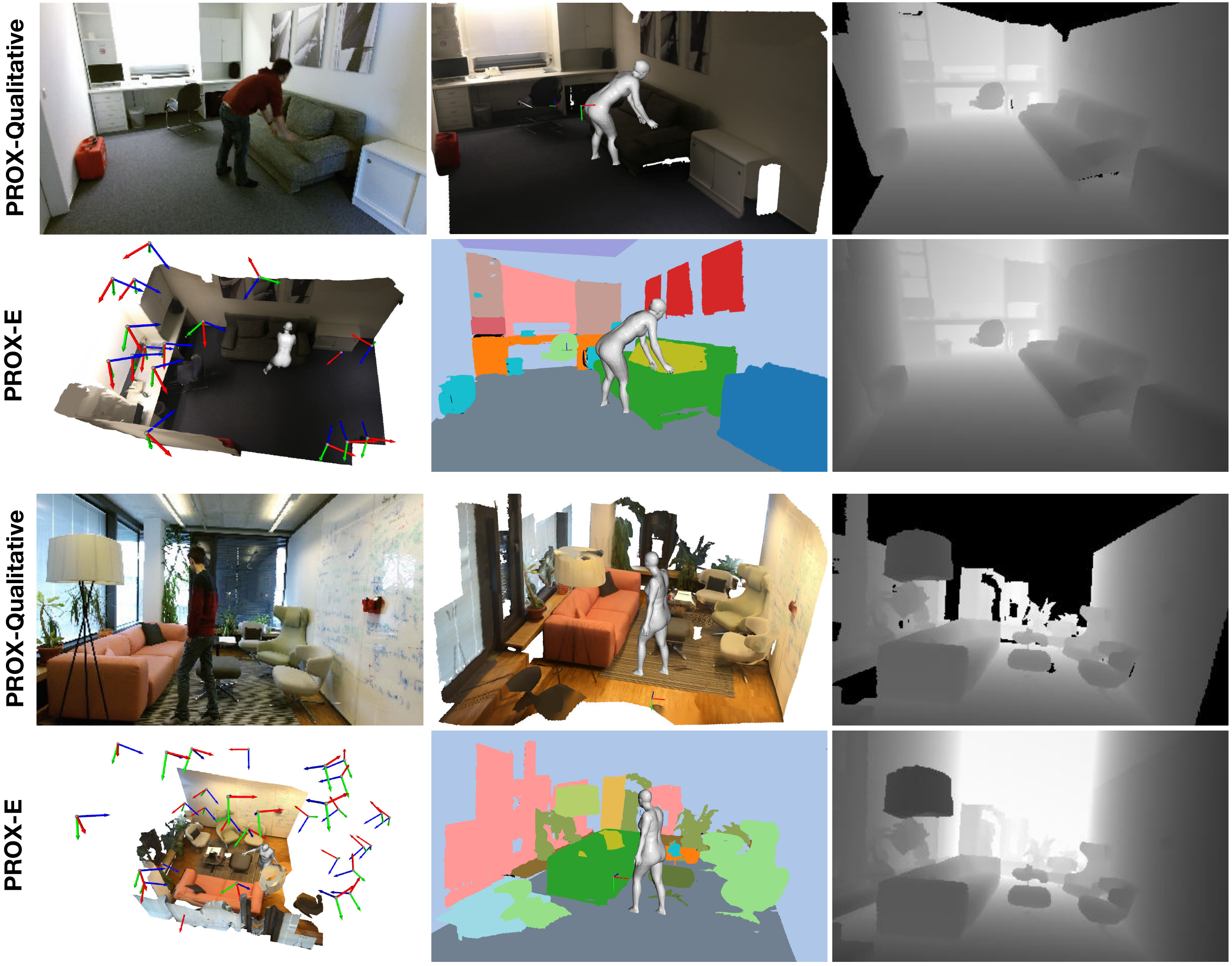}
    \vspace{-2em}
    \caption{Illustration of how we extend the {PROX-Qualitative} dataset \cite{PROX:2019} to {\bf PROX-E}. In the rows of {PROX-Qualitative}, a video frame, a body-scene mesh and a depth map are shown from left to right. In the rows of {\bf PROX-E}, the virtual camera setting, the mesh with semantics, and the completed depth map are shown from left to right.}
    \label{fig:prox_e}
\end{figure}

\myparagraph{PROX-E:} 
The {\bf PROX-E} dataset (pronounced ``proxy") is extended from the {PROX-Qualitative} (PROX-Q) dataset \cite{PROX:2019}, which records how people interact with various indoor environments. In {PROX-Q}, 3D human body meshes in individual frames are estimated by fitting the SMPL-X body model to the RGB-D data subject to scene constraints \cite{PROX:2019}. 
We use this data as pseudo-ground truth in our study, and 
extend {PROX-Q} in three ways: (1) We manually build up virtual walls, floors and ceilings to enclose the original open scans and simulate real indoor environments. (2) We manually annotate the mesh semantics following the object categorization of Matterport3D \cite{Matterport3D}. (3) We down-sample the original recordings and extract frames every 0.5 seconds. In each frame, we set up virtual cameras with various poses to capture scene depth and semantics. The optical axis of each virtual camera points towards the human body, and then Gaussian noise is applied on the camera translation. To avoid severe occlusion, all virtual cameras are located above half of the room height and below the virtual ceiling.
As a result, we obtain about 70K frames in total. We use `MPH16', `MPH1Library', `N0SittingBooth' and `N3OpenArea' as test scenes, and use samples from other scenes for training.
See Fig.~\ref{fig:prox_e}. 

\begin{figure}[]
    \centering
    \includegraphics[width=\linewidth]{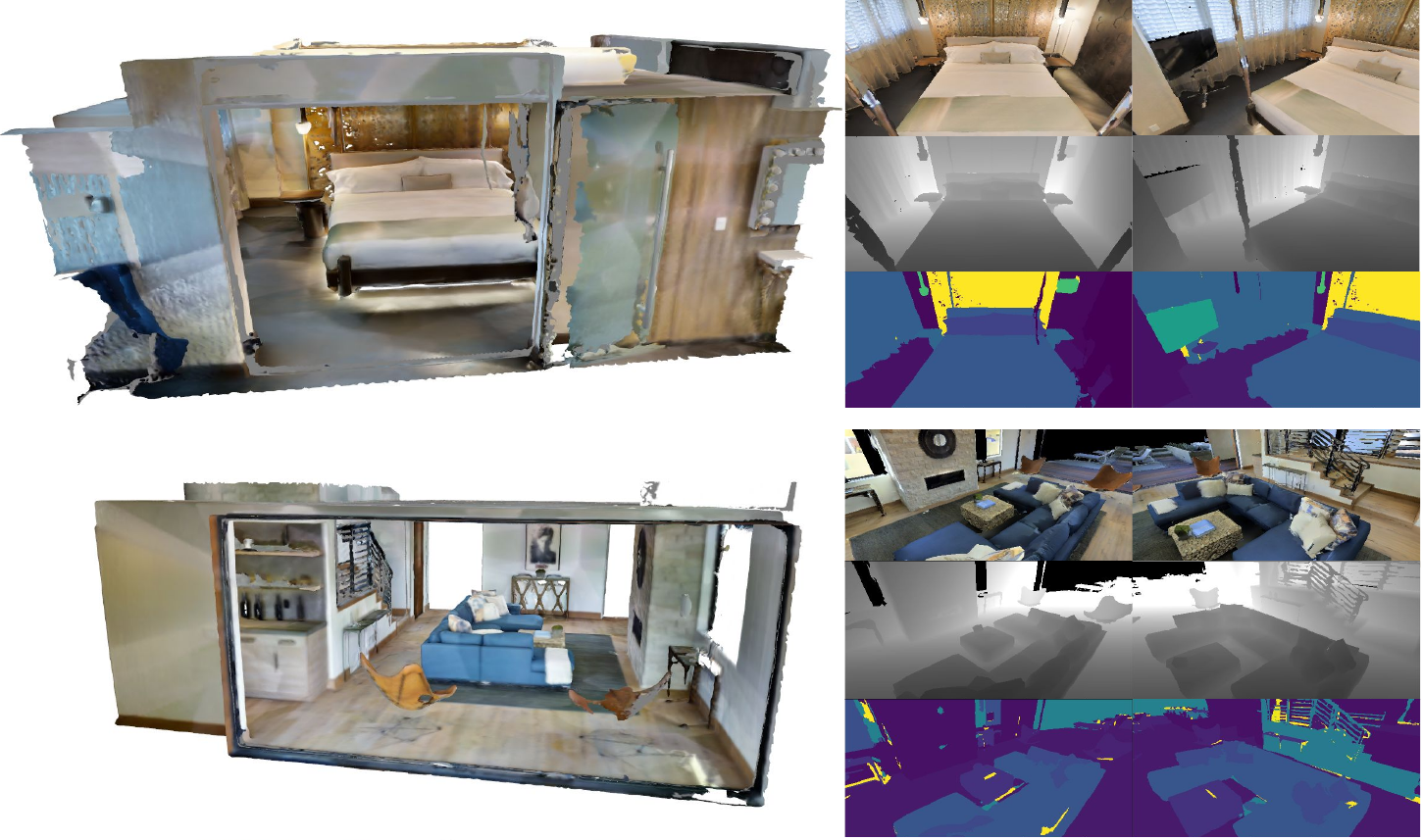}
    \caption{The left column shows two rooms in {\bf MP3D-R}. The right column shows snapshots captured by the Habitat virtual agent \cite{habitat19iccv} from different views, which contain RGB images, depth maps and scene semantics. }
    \label{fig:habitat_rooms}
\end{figure}

\myparagraph{MP3D-R:} This name denotes ``rooms in Matterport3D~\cite{Matterport3D}''.
From the architecture scans of Matterport3D, we extract 7 different rooms according to the annotated bounding boxes. In addition, we create a virtual agent using the Habitat simulator \cite{habitat19iccv}, and manipulate it to capture snapshots from various views in each room. We employ the RGB, the depth and the semantics sensor on the agent. These sensors are of height 1.8m from the ground, and look down at the scene; these are in a similar range as the virtual cameras in {\bf PROX-E}. For each snapshot, we also record the extrinsic and intrinsic parameters of the sensors. As a result, we obtain 32 snapshots in all 7 rooms.
Moreover, we follow the same procedure as in PROX-Qualitative \cite{PROX:2019} to calculate the SDF of the scene mesh. Our {\bf MP3D-R} is illustrated in Fig.~\ref{fig:habitat_rooms}. 

\subsubsection{Baseline} 
\label{sec:baseline}
To our knowledge, the most related work is Li et al.~\cite{li2019putting}, which proposes a generative model to put 3D body stick figures into images\footnote{The data and the pre-trained model in \cite{li2019putting} are based on SUNCG \cite{song2016ssc}, and not publicly available.}. For fair comparison, we modify their method to use SMPL-X to generate body meshes in 3D scenes. Specifically, we make the following modifications: (1) We change the scene representation from RGB (or RGB-D) to depth and semantics like ours to improve generalization. (2) During training, we perform K-means to cluster the VPoser pose features of training samples to generate the pose class. (3) The {\em where} module is used to infer the global translation, and the {\em what} module infers other SMPL-X parameters. (4) For training the geometry-aware discriminator, we project the body mesh vertices, rather than the stick figures, to the scene depth maps. We train the modified baseline model using {\bf PROX-E} with the default architecture and loss weights in ~\cite{li2019putting}. Moreover, we combine the modified baseline method with our scene geometry-aware fitting in our experiments.

\subsubsection{Evaluation: representation power}
Here we use {\bf PROX-E} to investigate how well the proposed network architectures represent human-scene interaction. We train all models using samples from virtual cameras in training scenes, validate them using samples from real cameras in training scenes, and test them using samples from real cameras in test scenes. For quantitative evaluation, we feed individual test samples to our models, and report the mean of the reconstruction errors, and the negative evidenced lower bound (ELBO), i.e.~$-log P(X)$, which is the sum of the reconstruction error and the KL divergence. For fair comparison, the reconstruction error of all models is based on $\mathcal{L}_{rec}$ in Eq. \ref{eq:rec}.
As shown in Tab.~\ref{tab:vae}, our models outperform the baseline model on both validation and test set by large margins. The metrics on the validation and the test sets are comparable, indicating that our virtual camera approach is effective in preventing severe over-fitting on the seen environments.

\begin{table}[t]
    \centering
    \footnotesize
    \caption{Comparison between models, in which ``+$\mathcal{L}_{HS}$'' denotes the model is trained with that human-scene interaction loss (Sec. \ref{sec:train_loss}). The best results are in boldface.} 
    \vspace{-1em}
    \begin{tabular}{lcccc}
    \toprule
    &     \multicolumn{2}{c}{rec. err.} & \multicolumn{2}{c}{$-log P(x)$}  \\
    \cmidrule(l{2pt}r{2pt}){2-3} \cmidrule(l{2pt}r{2pt}){4-5} 
    model & val & test & val & test\\
    \midrule
    baseline \cite{li2019putting}    & 0.52       & 0.48        & 0.98       & 0.72 \\
    S1                               & 0.22       & 0.25        & {\bf 0.23} & 0.41\\
    S1 + $\mathcal{L}_{HS}$          & {\bf 0.16} & 0.24        & 0.27       & {\bf 0.36} \\
    S2                               & 0.24       & 0.70        & 0.25       & 0.49 \\
    S2 + $\mathcal{L}_{HS}$          & 0.20       & {\bf 0.23}  & 0.30       & 0.39\\
    \bottomrule
    \end{tabular}
    \label{tab:vae}
\end{table}

\subsubsection{Evaluation: 3D body mesh generation}
Given a 3D scene, our goal is to generate diverse, physically and semantically plausible 3D human bodies. 
Based on \cite{li2019putting,wang2017binge}, we propose to quantitatively evaluate our method using a diversity metric and a physical metric. 
Also, we perform a user perceptual study to measure the semantic plausibility of the generated human bodies. 

The quantitative evaluation is based on the {\bf PROX-E} and the {\bf MP3D-R} dataset. 
When testing on {\bf PROX-E}, we train our models using all samples in the training scenes, and generate body meshes using the real camera snapshots in the testing scenes. 
For each individual model and each test scene, we randomly generate 1200 samples, and hence obtain 4800 samples.
When testing on {\bf MP3D-R}, we use all samples from {\bf PROX-E} to train the models. 
For each snapshot and each individual model, we randomly generate 200 samples, and hence obtain 6400 samples.

\myparagraph{(1) Diversity metric:} 
This metric aims to evaluate how diverse the generated human bodies are.
Specifically, we empirically perform K-means to cluster the SMPL-X parameters of all the generated human bodies to 20 clusters. 
Then, we compute the entropy (a.k.a Shannon index, a type of diversity index) of the cluster ID histogram of all the samples.
We also compute the average size of all the clusters.
A higher value indicates that the generated human bodies are more diverse in terms of their global positions, their body shapes and poses. 
We argue that this metric is essential for evaluating the quality of the generated bodies and should always be considered together with other metrics. 
For instance, a posterior-collapsed VAE, which always generates an identical body mesh, could lead to a low diversity score but superior performance according to the physical metric and the semantic metric. 

The results are shown in Tab.~\ref{tab:res_gen_diversity}. Overall, our methods consistently outperform the baseline. 
Notably, our methods increase the average cluster size of the generated samples by large margins, indicating that the generated human bodies are much more diverse than those from the baseline. 

\myparagraph{(2) Physical metric:} From the physical perspective, we evaluate the collision and the contact between the body mesh and the scene mesh. Given a scene SDF and a SMPL-X body mesh, we propose a non-collision score, which is calculated as the number of body mesh vertices with positive SDF values divided by the number of all body mesh vertices (10475 for SMPL-X). Simultaneously, if any body mesh vertex has a non-positive SDF value, then the body has contact with the scene. 
Then, for all generated body meshes, the non-collision score is the ratio of all body vertices in the free space, and the contact ratio is the calculated as the number of body meshes with contact divided by all generated body meshes. 
Therefore, due to the physical constraints, a higher non-collision score and contact ratio indicate a better generation, in analogy with precision and recall in an object detection task.

\begin{figure}
    \centering
    \includegraphics[width=\linewidth]{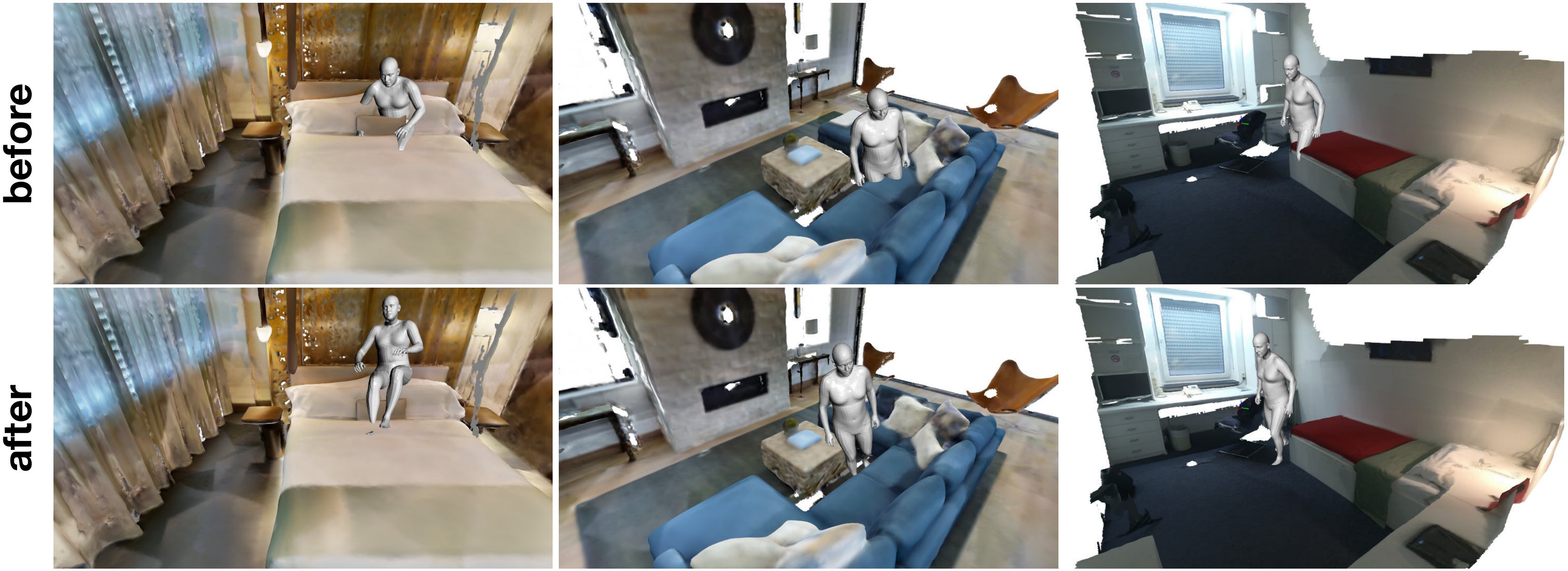}
    \vspace{-2em}
    \caption{Results before and after the scene geometry-aware fitting. }
    \label{fig:fitting}
\end{figure}

The results are presented in Tab.~\ref{tab:res_gen_physics}. 
First, one can see that our proposed methods consistently outperform the baseline for the physical metric. 
The influence of the $\mathcal{L}_{HS}$ loss on 3D body generation is not as obvious as on the interaction modeling task (see Tab.~\ref{tab:vae}).
Additionally, one can see that the scene geometry-aware fitting consistently improves the physical metric, since the fitting process aims to improve the physical plausibility. Fig.~\ref{fig:fitting} shows some generated examples before and after the fitting.

\begin{table}[t]
    \centering
    \footnotesize
    \caption{Comparison between different models according to the diversity metric. The best results for each metric are in boldface. ``S1'' and ``S2'' denote our stage-1 and stage-2 architecture, respectively. ``+ $\mathcal{L}_{HS}$'' denotes that the model is trained with the human-scene interaction loss (see Sec.~\ref{sec:train_loss}). ``+$\mathcal{L}_{f}$'' denotes the results are after the scene-aware fitting process (see Sec.~\ref{sec:fitting}). }
    \vspace{-1em}
    \begin{tabular}{lcccc}
    \toprule
     &  \multicolumn{2}{c}{cluster ID entropy} & \multicolumn{2}{c}{cluster size average} \\
     \cmidrule(lr){2-3} \cmidrule(lr){4-5}
    model & {\bf PROX-E} & {\bf MP3D-R} & {\bf PROX-E} & {\bf MP3D-R} \\

    \midrule
    baseline \cite{li2019putting}                 & 2.89      & 2.93       & 1.49         & 1.84 \\
    S1                                            & 2.96      & {\bf 2.99} & {\bf 2.51}   & 2.81 \\
    S1 + $\mathcal{L}_{HS}$                       & 2.93      & {\bf 2.99} & 2.40         & 2.73 \\
    S2                                            & {\bf 2.97}& 2.91       & 2.46         & 2.85 \\
    S2 + $\mathcal{L}_{HS}$                       & 2.96      & 2.89       & 2.22         & {\bf 2.90} \\
    \midrule
    baseline + $\mathcal{L}_{f}$                  & 2.93       & 2.92       & 1.52         & 1.94 \\
    S1 + $\mathcal{L}_{f}$                        & {\bf 2.97} & {\bf 2.98} & {\bf 2.53}   & 2.86 \\
    S1 + $\mathcal{L}_{HS}$ + $\mathcal{L}_{f}$   & 2.94       & 2.96       & 2.43         & 2.79 \\
    S2 + $\mathcal{L}_{f}$                        & 2.94       & 2.87       & 2.48         & 2.91 \\
    S2 + $\mathcal{L}_{HS}$ + $\mathcal{L}_{f}$   & 2.91       & 2.90       & 2.26         & {\bf 2.95} \\
    
    \bottomrule
    \end{tabular}
    \label{tab:res_gen_diversity}
\end{table}

\begin{table}[t]
    \centering
    \footnotesize
    \caption{Comparison between different models according to the physical metric. The best results are in boldface. }
    \vspace{-1em}
    \begin{tabular}{lcccc}
    \toprule
     &  \multicolumn{2}{c}{non-collision score} & \multicolumn{2}{c}{contact score} \\
     \cmidrule(lr){2-3} \cmidrule(lr){4-5}
    model & {\bf PROX-E} & {\bf MP3D-R} & {\bf PROX-E} & {\bf MP3D-R} \\

    \midrule
    baseline \cite{li2019putting}                 & 0.89       & 0.92       & 0.93       & 0.78 \\
    S1                                            & {\bf 0.93} & 0.94       & {\bf 0.95} & {\bf 0.80} \\
    S1 + $\mathcal{L}_{HS}$                       & 0.89       & {\bf 0.95} & 0.88       & 0.65 \\
    S2                                            & 0.91       & 0.93       & 0.88       & 0.79 \\
    S2 + $\mathcal{L}_{HS}$                       & 0.89       & {\bf 0.95} & 0.88       & 0.56 \\
    \midrule
    baseline + $\mathcal{L}_{f}$                  & 0.93       & 0.97       & {\bf 0.99} & {\bf 0.89} \\
    S1 + $\mathcal{L}_{f}$                        & {\bf 0.94} & 0.97       & {\bf 0.99} & 0.88 \\
    S1 + $\mathcal{L}_{HS}$ + $\mathcal{L}_{f}$   & 0.92       & {\bf 0.98} & {\bf 0.99} & 0.81 \\
    S2 + $\mathcal{L}_{f}$                        & {\bf 0.94} & 0.97       & {\bf 0.99} & 0.88 \\
    S2 + $\mathcal{L}_{HS}$ + $\mathcal{L}_{f}$   & 0.93       & 0.97       & {\bf 0.99} & 0.81 \\
    
    \bottomrule
    \end{tabular}
    \label{tab:res_gen_physics}
\end{table}

\myparagraph{(3) User study:}
In our study, we render our generated results as images, and upload them to Amazon Mechanical Turk (AMT) for a user study. Due to the superior performance of our S1 model without $\mathcal{L}_{HS}$, we compare it with the baseline, as well as ground truth if it exists. 
For each scene and each model, we generate 100 and 400 bodies in {\bf PROX-E} and {\bf MP3D-R}, respectively, and ask Turkers to give a score between 1 (strongly not natural) and 5 (strongly natural) to each individual result. The user study details are in the appendix.
Also, for each scene in the  {\bf PROX-E} dataset, we randomly select 100 frames from the ground truth \cite{PROX:2019}, and ask Turkers to evaluate them as well. 

The results are presented in Tab.~\ref{tab:user_study}. Not surprisingly, the ground-truth samples achieve the best score from the user study. We observe that the geometry-aware fitting improves the performance both for the baseline and our model, most likely due to the improvement of the physical plausibility.
Note that, although the baseline and our model achieve similar average scores, the diversity of our generated samples is much higher (Tab.~\ref{tab:res_gen_diversity}). This indicates that, compared to the baseline, our method generates more diverse 3D human bodies, while being equally good in terms of semantic plausibility given a 3D scene.

\begin{table}[t]
    \centering
    \footnotesize
    \caption{Comparison between models in the user study score (1-5). The best results for each metric are in boldface. }
    \vspace{-1em}
    \begin{tabular}{lcc}
    \toprule
    & \multicolumn{2}{c}{use study score w.r.t. mean$\pm$std}\\
    \cmidrule{2-3}
    model & {\bf PROX-E} & {\bf MP3D-R} \\
    \midrule
    baseline \cite{li2019putting}                 & 3.31 $\pm$ 1.39  & 3.14 $\pm$ 1.41 \\
    baseline + $\mathcal{L}_{f}$                  & 3.32 $\pm$ 1.35 & {\bf 3.35} $\pm$ 1.38 \\
    S1                                            & 3.29 $\pm$ 1.36 & 3.15 $\pm$ 1.40 \\
    S1 + $\mathcal{L}_{f}$                        & {\bf 3.49} $\pm$ {\bf 1.26} & {3.30} $\pm$ {\bf 1.30} \\
    \midrule
    ground truth                            & 4.04 $\pm$ 1.03 & n/a \\
    \bottomrule
    \end{tabular}
    \label{tab:user_study}
\end{table}

Qualitative results are shown in Fig.~\ref{fig:res_proxe}, Fig.~\ref{fig:res_mp3dr} and Fig.~\ref{fig:fail}. More results are in the appendix.

\begin{figure}
    \centering
    \includegraphics[width=\linewidth]{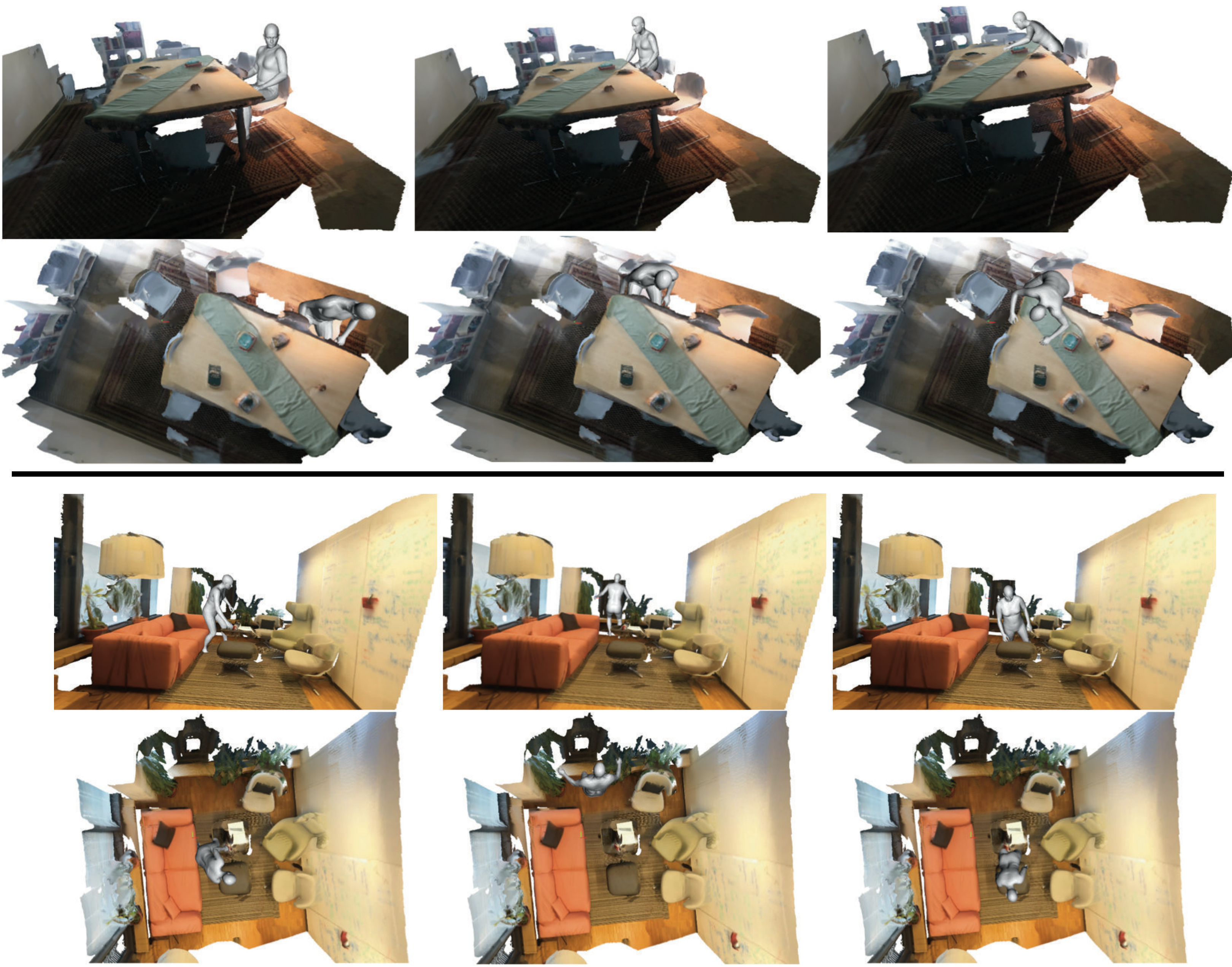}
    \vspace{-2em}
    \caption{Generated human bodies in two test scenes of {\bf PROX-E}. Results are visualized in two views.  }
    \label{fig:res_proxe}
\end{figure}

\begin{figure}
    \centering
    \includegraphics[width=\linewidth]{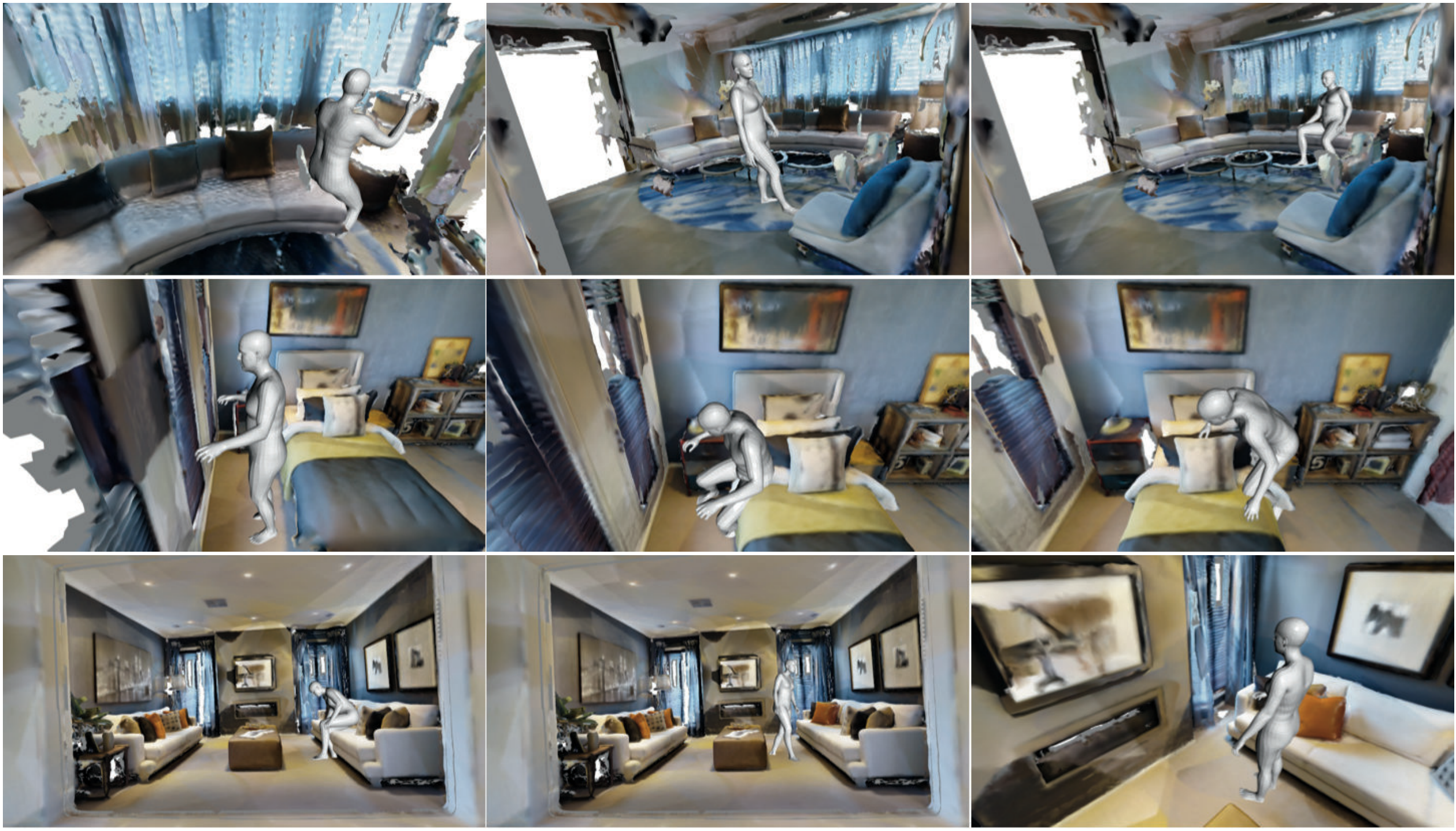}
    \vspace{-2em}
    \caption{Generated results in three scenes of {\bf MP3D-R}.} 
    \label{fig:res_mp3dr}
\end{figure}

\begin{figure}
    \centering
    \includegraphics[width=\linewidth]{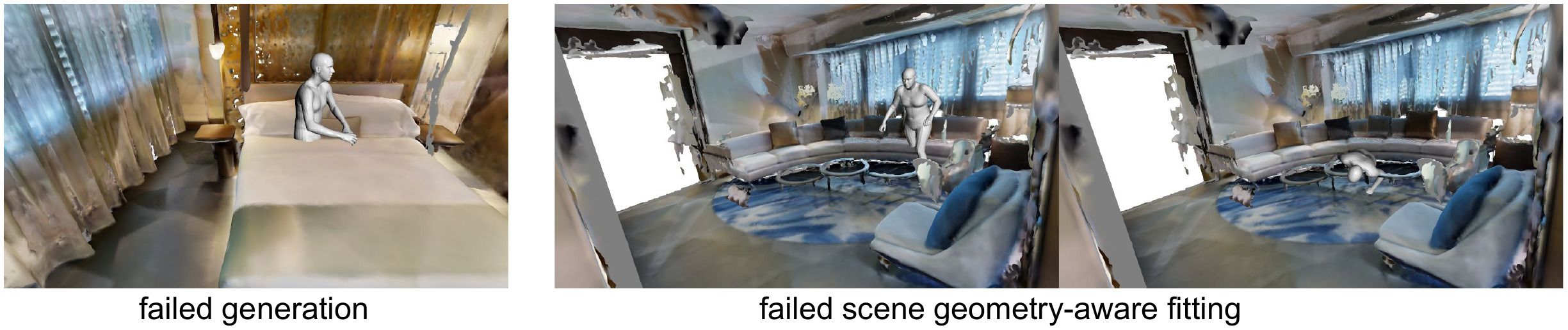}
    \vspace{-2em}
    \caption{Two typical failure cases in our results.}
    \label{fig:fail}
\end{figure}

\subsection{Scene-aware 3D Body Pose Estimation}
\label{sec:pose_estimation}
Here we perform a down-stream application and show our model improves 3D human pose estimation from monocular images. Given a RGB image of a scene without people, we estimate the depth map using the pre-trained model \cite{laina2016deeper}, and perform semantic segmentation using the model of \cite{chollet2017xception} pre-trained on the ADE20K \cite{zhou2017scene} dataset. To unify the semantics, we create a look-up table to convert the object IDs from ADE20K to Matterport3D. Next, we feed the estimated depth and semantics to our S1 model with $\mathcal{L}_{HS}$ 
and randomly generate 100 bodies. We compute the average of the pose features in the VPoser latent space, and denote it as $\theta_b^s$. 

When performing 3D pose estimation in the same scene, we follow the optimization framework of SMPlify-X \cite{SMPL-X:2019} and PROX \cite{PROX:2019}. In contrast to these two methods, we use our derived $\theta_b^s$ to initialize the optimization, and change the VPoser term in \cite[Eq.~7]{PROX:2019} from $|\theta_b|^2$ to $|\theta_b - \theta_b^s|^2$. We evaluate the performance using the {\bf PROX-Quantitative} dataset \cite{PROX:2019}. We derive the 2D keypoints from the frames via AlphaPose \cite{fang2017rmpe,xiu2018poseflow}, and obtain a $\theta_b^s$ from a background image without people. Then, we use the same optimization methods and the evaluation metric in \cite{PROX:2019} for fair comparison. The results are shown in Tab.~\ref{tab:pose_estimation}. 
We find that our method improves 3D pose estimation on the {\bf PROX-Quantitative} dataset.
This suggests that our model learns about the ways in which 3D people interact with 3D scenes.  Leveraging it as a scene-dependent body pose prior can improve 3D body pose estimation from RGB images.

\begin{table}
    \centering
    \footnotesize
    \caption{Results of 3D pose estimation from RGB frames in {\bf PROX-Quantitative}, in which ``PJE''/``p.PJE'' denote the mean per-joint error without/with Procrustes alignment, and ``V2V''/``p.V2V'' denote the mean vertex-to-vertex error without/with Procrustes alignment, respectively.}
    \begin{tabular}{lcccc}
    \toprule
         & \multicolumn{4}{c}{Error (in millimeters)} \\
         \cmidrule{2-5}
     method & PJE & V2V & p.PJE & p.V2V \\
     \midrule
     Simplify-X \cite{SMPL-X:2019} & 223.83 & 225.60 & 73.28 & 62.93 \\
     PROX \cite{PROX:2019}         & {\bf 171.78} & 173.97 & 73.20 & 64.76 \\
     Ours                          & {174.10} & {\bf 171.75} & {\bf 71.73} & {\bf 62.64}\\
    \bottomrule
    \end{tabular}
    \label{tab:pose_estimation}
\end{table}

%% file: latex/sec5-conclusion.tex
\section{Conclusion}

In this work, we introduce a generative framework to produce 3D human bodies that are posed naturally in the 3D environment. 
Our method consists of two steps: (1) A scene context-aware human body generator is proposed to learn a distribution of 3D human pose and shape, conditioned on the scene depth and semantics; (2) geometry-aware fitting is employed to impose physical plausibility of the human-scene interaction. Our experiments demonstrate that the automatically synthesized 3D human bodies are realistic and expressive, and interact with 3D environment in a semantic and physical plausible way.

\myparagraph{Acknowledgments.} We sincerely acknowledge: Joachim Tesch for all his work on graphics supports. Xueting Li for implementation advice about the work \cite{li2019putting}. David Hoffmann, Jinlong Yang, Vasileios Choutas, Ahmed Osman, Nima Ghorbani and Dimitrios Tzionas for insightful discussions. Daniel Scharstein and Cornelia K\"{o}hler for proof reading. Benjamin Pellkofer and Mason Landry for IT/hardware supports. 
Y. Z. and S. T. acknowledge funding by Deutsche Forschungsgemeinschaft (DFG, German Research Foundation) Projektnummer 276693517 SFB 1233.

\myparagraph{Disclosure.} MJB has received research gift funds from Intel, Nvidia, Adobe, Facebook, and Amazon. While MJB is a part-time employee of Amazon, his research was performed solely at MPI. He is also an investor in Meshcapde GmbH.